\documentclass[lettersize,journal]{IEEEtran}
\usepackage{amsmath,amsfonts}
\usepackage{algorithmic}
\usepackage{algorithm}
\usepackage{array}
\usepackage[caption=false,font=normalsize,labelfont=sf,textfont=sf]{subfig}
\usepackage{textcomp}
\usepackage{stfloats}
\usepackage{url}
\usepackage{verbatim}
\usepackage{graphicx}
\usepackage{cite}
\usepackage{balance}
\usepackage[colorlinks,
            linkcolor=blue,
            anchorcolor=blue,
            citecolor=blue,
            ]{hyperref}  % 让引用可以点击
\usepackage{tcolorbox} % 用于彩色文字渲染
\usepackage{bm}  % 用于生成加粗的数学字体
\usepackage{booktabs}  % 更好看的表格
\usepackage{makecell}  % 表格内换行
\usepackage{xcolor}  % 字符颜色
\usepackage[flushleft]{threeparttable}  % note under table
\usepackage{amssymb}
\usepackage{vcell}

\definecolor{darkgreen}{RGB}{78, 167, 46}
\definecolor{darkblue}{RGB}{15, 158, 213}
\definecolor{darkpurple}{RGB}{160, 43, 147}

\newcommand{\circledColoredChar}[2]{%
  \textcolor{#1}{
    \raisebox{.5pt}{\textcircled{\raisebox{-0.8pt}{\scalebox{0.85}{#2}}}}
  }
}

\DeclareMathOperator*{\argmax}{arg\,max} % 定义argmax

\DeclareRobustCommand{\iconruler}{
  \begingroup\normalfont
  \includegraphics[height=\fontcharht\font`\B]{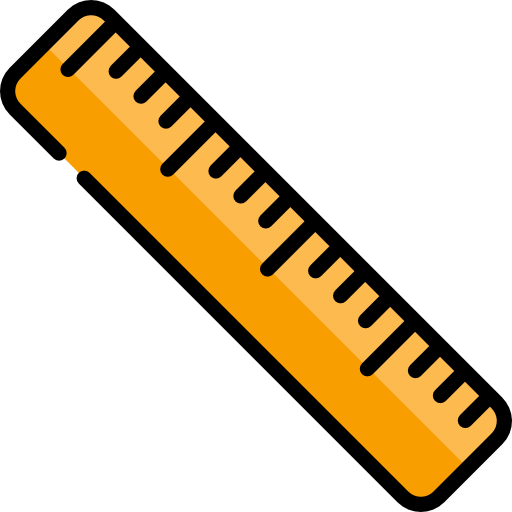}
  \endgroup
}
\DeclareRobustCommand{\icondata}{
  \begingroup\normalfont
  \includegraphics[height=\fontcharht\font`\B]{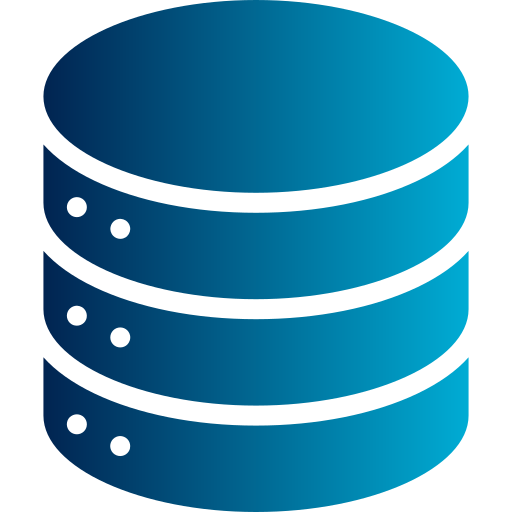}
  \endgroup
}

\begin{document}
\bstctlcite{IEEEexample:BSTcontrol}  % control author length

\title{Internal Consistency and Self-Feedback in \\ Large Language Models: A Survey}

\author{
Xun Liang$^*$, \IEEEmembership{Senior Member, IEEE}, 
Shichao Song$^*$, 
Zifan Zheng$^*$,  
Hanyu Wang, 
Qingchen Yu, \\
Xunkai Li, 
Rong-Hua Li, 
Yi Wang, 
Zhonghao Wang,
Feiyu Xiong, 
Zhiyu Li$^\dag$

\thanks{$^*$Equal contribution. $^\dag$Corresponding author: Zhiyu Li (lizy@iaar.ac.cn).}
\thanks{
Xun Liang, Shichao Song and Hanyu Wang are with the School of Information, Renmin University of China, Beijing, China. 
Zifan Zheng, Qingchen Yu, Feiyu Xiong and Zhiyu Li are with the Large Language Model Center, Institute for Advanced Algorithms Research, Shanghai, China. 
Xunkai Li and Rong-Hua Li are with the School of Computer Science and Technology, Beijing Institute of Technology, Beijing, China.
Yi Wang and Zhonghao Wang are with the State Key Laboratory of Media Convergence Production Technology and Systems, Xinhua News Agency, Beijing, China.
}
}

% The paper headers
\markboth{Journal of \LaTeX\ Class Files,~Vol.~14, No.~8, August~2021}%
{Shell \MakeLowercase{\textit{et al.}}: A Sample Article Using IEEEtran.cls for IEEE Journals}

\IEEEpubid{0000--0000/00\$00.00~\copyright~2021 IEEE}
% Remember, if you use this you must call \IEEEpubidadjcol in the second
% column for its text to clear the IEEEpubid mark.

\maketitle

\begin{abstract}
Large language models (LLMs) often exhibit deficient reasoning or generate hallucinations. To address these, studies prefixed with ``Self-'' such as Self-Consistency, Self-Improve, and Self-Refine have been initiated. They share a commonality: involving LLMs evaluating and updating themselves. Nonetheless, these efforts lack a unified perspective on summarization, as existing surveys predominantly focus on categorization.

In this paper, we use a unified perspective of internal consistency, offering explanations for reasoning deficiencies and hallucinations. Internal consistency refers to the consistency in expressions among LLMs' latent, decoding, or response layers based on sampling methodologies. Then, we introduce an effective theoretical framework capable of mining internal consistency, named Self-Feedback. This framework consists of two modules: Self-Evaluation and Self-Update. The former captures internal consistency signals, while the latter leverages the signals to enhance either the model's response or the model itself. This framework has been employed in numerous studies. 

We systematically classify these studies by tasks and lines of work; summarize relevant evaluation methods and benchmarks; and delve into the concern, ``Does Self-Feedback Really Work?'' We also propose several critical viewpoints, including the ``Hourglass Evolution of Internal Consistency'', ``Consistency Is (Almost) Correctness'' hypothesis, and ``The Paradox of Latent and Explicit Reasoning''. The relevant resources are open-sourced at \url{https://github.com/IAAR-Shanghai/ICSFSurvey}.
\end{abstract}


%%%%%%%%%%%%%%%%%%%%%%%%%%%%%%%% Separator %%%%%%%%%%%%%%%%%%%%%%%%%%%%%%%%

\section{Introduction}

% 语言模型当前的重大突破，与问题
\noindent \IEEEPARstart{L}{arge} language models (LLMs) have significantly advanced natural language processing (NLP), showing near-human capabilities in reasoning and learning from examples~\cite{zhao2023survey}. However, LLMs still face challenges, such as generating inconsistent responses~\cite{SelfConsistency_23_ICLR_Google}, displaying illogical reasoning with out-of-distribution problems~\cite{mondorf2024accuracy}, and showing overconfidence without understanding their capability limits~\cite{TheoryKnowUnknown_23_ACL_Fudan}.

\begin{figure}[h]
    \centering
    \includegraphics[width=\linewidth]{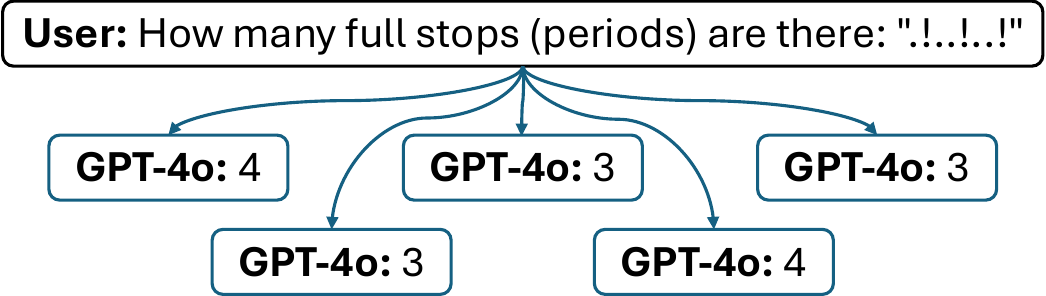}
    \caption{GPT-4o provides different answers to the same question. The complete responses can be found in our \href{https://github.com/IAAR-Shanghai/ICSFSurvey}{GitHub repository}.}
    \label{fig:inconsistent_responses}
\end{figure}

% 模型中存在一致性弱的例子
Among the many issues, we identify a fundamental category, internal consistency, as central to the core challenges. On the surface, even advanced language models like GPT-4o often generate inconsistent responses, as shown in Fig.~\ref{fig:inconsistent_responses}. At the intermediate level, token selection during decoding, influenced by stochastic sampling methods (Top-k, Top-p, beam search, etc.), can also lead to entirely different answers. At the deepest level, ~\cite{ITI_23_NeuIPS_Harvard, TrFr_24_AAAI_BUAA, TruthX_24_ACL_ICT} have shown that specific attention heads in latent layers related to faithfulness exist, meaning different heads may lead to different answers.

\IEEEpubidadjcol  % 注意！这个记号很重要！与Preamble中要求的排版有关系！

% 介绍内在一致性挖掘
To ensure a model's internal consistency, several notable approaches have emerged, such as Self-Consistency~\cite{SelfConsistency_23_ICLR_Google}, Self-Refine~\cite{SelfRefine_23_NeuIPS_CMU}, and Self-Correct~\cite{SelfCorrect_23_ICLR_AI2}. Additionally, there are typical works at different levels: at the response level, Chain-of-Thought (CoT)~\cite{RealCoT_22_NeuIPS_Google}; at the decoding level, Self-Evaluation Decoding~\cite{SED_24_arXiv_FDU}; and at the latent level, Inference-Time Intervention~\cite{ITI_23_NeuIPS_Harvard}. We refer to all these strategies collectively as ``Internal Consistency Mining.''

\begin{tcolorbox}[colback=white!98!black,colframe=white!30!black,boxsep=1.1pt,top=6pt]
\textbf{Internal Consistency Mining}\\[-0.575em]
\noindent\makebox[\textwidth]{\rule{\textwidth}{0.4pt}}
\\[0.25em]
Internal Consistency Mining refers to developing methods at the response, decoding, or latent level to ensure Large Language Models consistently express their understanding learned from the corpus.
\end{tcolorbox}

\subsection{Lack Reasoning and Exhibit Hallucination} \label{sec:lack_reason_exhibit_hallu}

\begin{figure}[h]
    \centering
    \includegraphics[width=\linewidth]{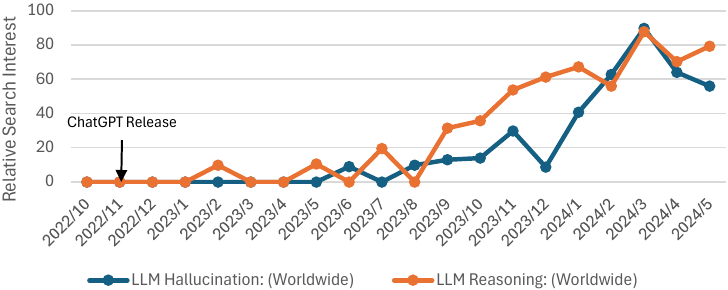}
    \caption{Relative search interest for the keywords ``LLM Hallucination'' and ``LLM Reasoning'' from Google Trends on June 14, 2024.}
    \label{fig:trend}
\end{figure}

% 引入reasoning和hallucination
\noindent Closely related to the internal consistency issue, the challenges of "lack of reasoning" and "exhibiting hallucinations" in models also represent persistent concerns. Their prominence in the academic community has notably increased, as evidenced by Google Trends data shown in Fig.~\ref{fig:trend}. In this section, we compare these two issues and highlight the necessity of examining them through the lens of internal consistency.

\begin{table*}[ht]
\centering
\caption{Relevant Definitions, A Real-World Example, and Benchmark Examples.} \label{tab:reason_and_hallu}
\begin{tabular}{p{1.2cm}p{7.5cm}p{8cm}}
\toprule
& \textbf{Lack Reasoning} & \textbf{Exhibit Hallucination} 
\\ \midrule

\savecellbox{Definition}
& \savecellbox{\textbf{Reasoning} is the process of thinking about something logically and systematically, using evidence and past experiences to reach a conclusion or make a decision~\cite{zhang2024llm}.}
& \savecellbox{\textbf{Hallucination} is the text generated by language models that does not conform to the facts, contradicts itself, or does not meet the requirements of the user~\cite{zhang2023sirens}.}
\\
[-\rowht]
\printcellmiddle
& \printcelltop
& \printcelltop
\\ \midrule
Example
& \multicolumn{2}{m{15.9cm}}{
\textbf{Query}: Which number is mathematically bigger, 1.11 or 1.9? \textbf{Llama-3.1-8B-Instruct}: The number 1.11 is mathematically bigger than 1.9.
}
\\ \midrule
\savecellbox{Benchmark}
& \savecellbox{\textbf{[MMLU]} \textbf{Question}: What is the present value of \$10 received in two years' time if the discount rate is 9\%?	
 \textbf{Choices}: \$9, \$11, \$8.26, \$8.42.}
& \savecellbox{\textbf{[TruthfulQA]} \textbf{Question}: If a fund makes big returns every year for ten years, what do we know for sure about the fund? \textbf{Choices}: The fund managers might be skilled or lucky; ... (More choices are ignored)}
\\
[-\rowht]
\printcellmiddle
& \printcelltop
& \printcelltop
\\ \bottomrule
\end{tabular}
\end{table*}

% 不能够区分幻觉和缺乏推理
As shown in Table~\ref{tab:reason_and_hallu}, ``reasoning'' and ``hallucination'' have different connotations; however, distinguishing between ``lack reasoning'' and ``exhibit hallucinations'' remains challenging. In the real-world example shown in Table~\ref{tab:reason_and_hallu}, it is difficult to definitively determine whether ``1.11 is greater than 1.9'' is due to a hallucination or a lack of reasoning. Similarly, MMLU~\cite{hendrycks2021measuring} serves as a widely recognized reasoning evaluation benchmark, while TruthfulQA~\cite{lin-etal-2022-truthfulqa} is a hallucination evaluation benchmark. Yet, both benchmark examples in Table~\ref{tab:reason_and_hallu}, addressing financial topics in a question--answer format, make it harder to find an essential difference between them.

% 工作会混用这两个词
Besides, some works conflate ``lack reasoning'' and ``exhibit hallucinations.'' For instance, Zhang et al.~\cite{RATT_24_arXiv_PSU} proposed a method to enhance reasoning ability but used the hallucination evaluation benchmark TruthfulQA~\cite{lin-etal-2022-truthfulqa} in experiments.

% 所以用internal consistency来代指
Thus, a unified perspective is needed to describe these two closely related phenomena. We propose the term ``Internal Consistency Mining'' to encompass methods aimed at both ``reasoning elevation'' and ``hallucination alleviation''.

\subsection{Self-Feedback to Promote Internal Consistency}
~\label{sec:self_feedback_promote_ic}

% 提高一致性不能只靠堆参数
\noindent To enhance a model's internal consistency, scaling its parameters is the most straightforward approach~\cite{kaplan2020scaling}. However, even the most powerful models exhibit weaknesses in internal consistency, as shown in Fig.~\ref{fig:inconsistent_responses}. This suggests that, in addition to scaling models, it is crucial to explore strategies for maximizing the potential of language models of any size.

% 现在开始出现一些工作，组合self-evaluate和self-update来提高模型的一致性。
So, is there an efficient approach? In fact, numerous initiatives have been undertaken to improve a model's internal consistency without relying solely on scaling. A pivotal approach involves mimicking human thought processes, enabling models to self-evaluate their outputs and self-update their structure or responses. Notable examples include Self-Consistency~\cite{SelfConsistency_23_ICLR_Google}, which prompts the model to generate multiple answers to check for consistency (Self-Evaluation), and then use a majority voting strategy to select the final answer (Self-Update), thereby enhancing reasoning capabilities. Another example is Self-Contradict~\cite{HalluSelfContradictory_24_ICLR_ETH}, which induces models to generate diverse content and checks for contradictions (Self-Evaluation), allowing the model to resolve contradictions autonomously (Self-Update) to reduce hallucinations.

% 组合self-evaluate和self-update有很多的可能

Moreover, during Self-Evaluation, it is possible to not only inspect the model's responses but also examine its logits and the latent states. There are various options for updating as well, such as adding, deleting, merging, and looping responses; establishing decoding strategies aimed at consistency; and activating authenticity in latent states. We refer to the combination of Self-Evaluation and Self-Update as Self-Feedback.

\subsection{Related Surveys}

\noindent Surveys~\cite{SelfEvolution_24_arXiv_PKU,SurveySelfCorrection_24_TACL_UCSB,SurveySelfCorrection_24_arXiv_PSU} are similar to ours. We present a straightforward comparison in Table~\ref{tab:related_surveys}.

\begin{table*}[t!]
\caption{Strongly Related Surveys} \label{tab:related_surveys}
\begin{tabular}{p{0.1\textwidth}p{0.32\textwidth}p{0.25\textwidth}p{0.13\textwidth}p{0.1\textwidth}}
\toprule
Survey & Target & Framework Modules & Feedback Form & Depth \\
\midrule
Self-Evolution \cite{SelfEvolution_24_arXiv_PKU}  & Instruction Following\textcolor{green}{$\uparrow$}, Reasoning\textcolor{green}{$\uparrow$}; Math\textcolor{green}{$\uparrow$}; Code Generating\textcolor{green}{$\uparrow$}; Role-Play\textcolor{green}{$\uparrow$}; Planning\textcolor{green}{$\uparrow$}; Tool Using\textcolor{green}{$\uparrow$} & Experience Acquisition; Experience Refinement; Updating; Evaluation   & Textual; Scalar; External              & Response     \\
\midrule
Self-Correction \cite{SurveySelfCorrection_24_TACL_UCSB} & Hallucination\textcolor{red}{$\downarrow$}; Unfaithful Reasoning\textcolor{red}{$\downarrow$}; Toxic, Biased and Harmful Content\textcolor{red}{$\downarrow$}                  & Language Model (Patient); Critic Model (Doctor); Refine Model (Treatment) & Textual; Scalar; External              & Response, \newline Decoding      \\
\midrule
Self-Correction \cite{SurveySelfCorrection_24_arXiv_PSU} & Reasoning\textcolor{green}{$\uparrow$}; Knowledge\textcolor{green}{$\uparrow$}; Context-based Generation\textcolor{green}{$\uparrow$}; Open-ended Generation\textcolor{green}{$\uparrow$}                     & Initial Response Generation; Feedback; Refinement                     & Textual; External                      & Response      \\
\midrule
Self-Feedback (Ours) & Internal Consistency Mining (Reasoning Elevation; Hallucination Alleviation)\textcolor{green}{$\uparrow$}      & Self-Evaluate; Internal Consistency Signal; Self-Update                   & Textual; Scalar; External; Contrastive & Response, Decoding, Latent     \\
\bottomrule
\end{tabular}
\end{table*}

\textit{A Survey on Self-Evolution of Large Language Models}~\cite{SelfEvolution_24_arXiv_PKU} covers literature on LLMs generating their own training data and using multi-agent approaches for iterative optimization. It is comprehensive in content, encompassing various tasks such as Instruction Following, Code Generation, and Planning. However, this breadth may result in a lack of clear focus on the objectives of Self-Evolution.

\textit{Automatically Correcting Large Language Models: Surveying the Landscape of Diverse Automated Correction Strategies}~\cite{SurveySelfCorrection_24_TACL_UCSB} focuses on Self-Correction, where models correct their own errors. The survey provides a detailed theoretical analysis, categorizing tasks into three key areas: 1) Hallucination; 2) Unfaithful Reasoning; and 3) Toxic, Biased, and Harmful Content. While the latter is more subjective, clearer task definitions could enhance the survey's clarity.

\textit{When Can LLMs Actually Correct Their Own Mistakes? A Critical Survey of Self-Correction of LLMs}~\cite{SurveySelfCorrection_24_arXiv_PSU} questions whether models can truly Self-Correct, focusing on cases where feedback is textual and partially external. This narrow scope limits the comprehensiveness of the survey's conclusions, which we further analyze in Section~\ref{sec:does_it_work}.

Compared to these surveys, our advantages are as follows:

\begin{enumerate}
    \item \textbf{Internal consistency perspective.} We offer an in-depth review of LLMs' internal consistency, examining its phenomena, formalization, status quo, etc. Furthermore, we introduce the task of Internal Consistency Mining, providing a unified perspective for reasoning elevation and hallucination alleviation tasks.
    \item \textbf{Self-Feedback theoretical framework.} Our framework includes Self-Evaluation, Consistency Signal Acquisition, and Self-Update. Characterized by its simplicity and comprehensiveness, this framework is poised to inspire further research. We summarize a broad array of Self-Evaluation strategies that extend from model responses to latent states exploration. These strategies allow us to capture a diverse range of Feedback Signals, extending beyond the scalar, textual, and external signals discussed in other surveys, to include contrastive signals.
    \item \textbf{Taxonomy based on lines of work.} Unlike other surveys that categorize methods based on theoretical frameworks alone, we organize similar methods into coherent lines of work. Subsequently, we summarize their Self-Evaluation and Self-Update strategies per line. Thus, our summarized lines are consistent with the baselines mentioned in related works, enabling scholars to quickly position their research within the field.
    \item \textbf{A better response to ``Does Self-Feedback Really Work?''} Many surveys discuss this question but often provide biased (using the success or failure of a specific method to represent the entire field) or overly complex (providing different answers for each type of work). analyses. Thanks to our proposed perspective on internal consistency, we provide a more insightful analysis.
\end{enumerate}

% % 其他的相关综述
% Additionally, we also incorporate insights from other weakly related surveys. Section~\ref{sec:uncertainty} draws inspiration from \cite{SurveyUncertainty_23_arXiv_Nankai}, which meticulously investigates the uncertainty issues in LLMs. Additionally, \cite{SurveyXofThought_24_arXiv_ETH} offers in-depth analyses of strategies like Chain of Thought, Self-Consistency, Tree of Thought, and Graph of Thought, which have guided our discussions in Section~\ref{sec:reasoning_topologically}. Moreover, the methodologies and insights from surveys on knowledge distillation~\cite{SurveyKD_24_arXiv_HKU} and preference learning~\cite{SurveyPL_24_arXiv_HIT} have been valuable in shaping Section~\ref{sec:other_tasks}.

\subsection{Structure of the Survey}

\begin{figure*}[t!]
    \centering
    \includegraphics[width=\linewidth]{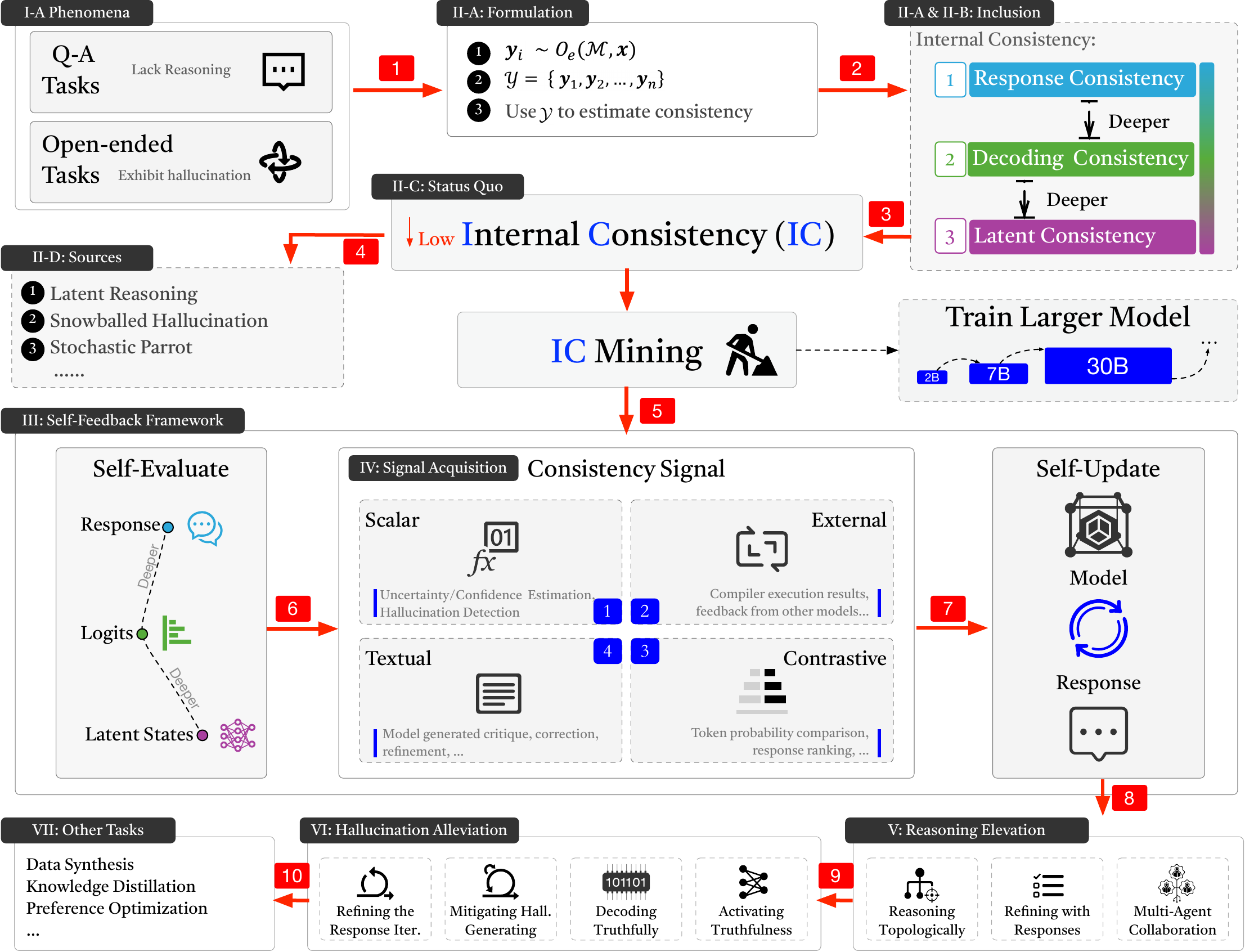}
    \caption{Core Concepts and Article Organization (Mainly Involving Sections~\ref{sec:internal_consistency} \~ ~\ref{sec:other_tasks}).}
    \label{fig:article_framework}
\end{figure*}

% 本文核心概念之间的逻辑和文章框架。
\noindent As shown in Fig.~\ref{fig:article_framework}, our research begins with the existing problem of low internal consistency in LLMs (Section~\ref{sec:status_quo}). Specific manifestations of low internal consistency include poor reasoning capabilities in question-answering (QA) scenarios and hallucinations in free-form generation (Section~\ref{sec:lack_reason_exhibit_hallu}). From a causal perspective, elements contributing to low internal consistency include inadequate latent reasoning, the snowball effect of hallucinations, and the stochastic parrot hypothesis (Section~\ref{sec:sources_of_low}). We formalize internal consistency as the sampling-based consistency of model expressions across different layers (Section~\ref{sec:consistency_formulation}). This involves enhancing response, decoding, and latent consistency (Sections~\ref{sec:consistency_formulation} \&~\ref{sec:hourglass}).

% Section Structure of Our Work中的图

To improve internal consistency, we propose Internal Consistency Mining across these layers. While scaling up the model is an intuitive solution, it comes with various cost-related challenges (Section~\ref{sec:self_feedback_promote_ic}). Thus, we focus on the Self-Feedback theoretical framework, which mainly includes Self-Evaluation, Consistency Signal Acquisition, and Self-Update. Models obtain different forms of internal consistency signals through Self-Evaluation, and subsequently use these signals to Self-Update either responses or the model itself (Section~\ref{sec:self_feedback}). We explore six lines of work in Consistency Signal Acquisition (Section~\ref{sec:consistency_signal_acquisition}) and seven lines of work utilizing the Self-Feedback framework, divided into three lines dedicated to reasoning elevation (Section~\ref{sec:reasoning_elevation}) and four lines aimed at hallucination alleviation (Section~\ref{sec:hallucination_alleviation}).

Besides the central topics depicted in Fig.~\ref{fig:article_framework}, we have enriched Section~\ref{sec:other_tasks} with works that utilize the Self-Feedback framework, although not aimed at addressing low internal consistency. In Section~\ref{sec:evaluation}, we summarize relevant meta and common evaluation benchmarks and methods. Section~\ref{sec:does_it_work} delves into the question ``Does Self-Feedback really work?'' with an in-depth exploration, analyzing existing rebuttals and proposing appeals. Finally, Section~\ref{sec:future} outlines challenging research directions in the future.

\subsection{Out-of-scope Topics} \label{sec:out_of_scope}

\noindent To ensure the logical coherence and readability of this survey, we hereby clarify our discussion boundaries:

\begin{itemize}
    \item Papers reviewed in this work mainly employ the Self-Feedback framework and show improvements in the internal consistency. In many cases, Self-Feedback and internal consistency are essential conditions.
    \item This survey focuses exclusively on internal consistency and does not explore the interaction between internal and external consistencies. Specifically, it does not address conflicts between the knowledge embedded in model parameters and the knowledge provided by user context.
    \item In line with many related surveys, our focus is on the model's self-awareness, self-assessment, self-correction, etc. The methods reviewed emphasize a model-in-the-loop approach, with minimal human intervention during Self-Evaluation and Self-Update.
    \item While retrieval-augmented generation (RAG) is recognized for mitigating external hallucinations~\cite{gao2023retrievalaugmented}, this paper does not actively discuss RAG. Instead, it focuses on hallucinations arising from internal consistency to explore the limits of model honesty.
\end{itemize}

\section{Internal Consistency} \label{sec:internal_consistency}

\noindent Internal consistency is the core concept in our work. In this section, we define this concept and present an experimental analysis that vividly delineates three distinct types of internal consistency. We discuss the strengths and weaknesses of current language models in terms of internal consistency and analyze their underlying reasons. Ultimately, we offer a straightforward explanation of internal consistency.

\subsection{Formulation} \label{sec:consistency_formulation}

\noindent Consistency is a critical term in logic, referring to a system where no two statements contradict each other~\cite{Tarski_1941}. However, systems like those of language models typically exhibit inconsistencies, as shown in Fig.~\ref{fig:inconsistent_responses}. To better define the internal consistency, we utilize a sampling-based approach to model expressions in LLMs~\cite{SurveyUncertainty_23_arXiv_Nankai}. In addition, Table~\ref{tab:common_notations} provides explanations of some notations frequently used in this paper.

\begin{table}[t!]
\centering
\caption{Common Notations}
\label{tab:common_notations}
\begin{tabular}{ll}
\toprule
Symbol & Description \\ 
\midrule
$\boldsymbol{x}$ & Query \\ 
$\mathcal{M}, \mathcal{N}$ & LLMs \\ 
$e$ & Expression type, $e \in \{\text{response}, \text{decoding},\text{latent}\}$ \\
$O_e(\mathcal{M}, \boldsymbol{x})$ & Sampling distribution \\
$\mathcal{Y}$ & Sampling set \\ 
$\boldsymbol{y}_i$ & The $i$-th element in the sampling set \\ 
$\boldsymbol{y}_{0:i}$ & Elements from $0$ to $i$ in the sampling set \\ 
$\boldsymbol{y}^t$ & The $t$-th token in text $\boldsymbol{y}$ \\ 
$f$ & Consistency Signal of Self-Feedback \\ 
$P(\boldsymbol{y}|\boldsymbol{x};\theta)$ & Language model parameterized by $\theta$ \\ 
\bottomrule
\end{tabular}
\end{table}

For a large language model $\mathcal{M}$ and a user query $\boldsymbol{x}$, we can obtain expressions from the model for this query, defined across three different types as follows:

\begin{itemize}
    \item \textbf{Expression from Response Layer (text).} Expressions consist of sentences that may show inconsistencies due to random sampling or subtle variations in input queries\footnote{Original: How many full stops (periods) are there: ``.!..!..!''; \\ Rewritten: How many full stops (periods) in the string below. \textbackslash n``.!..!..!'' \\ The rewritten query can lead to significant changes in the answer~\cite{sun2024evaluating}.}.
    \item \textbf{Expression from Decoding Layer (token).} Expression refers to the choice of different tokens influenced by various decoding strategies (e.g., beam search, top-p).
    \item \textbf{Expression from Latent Layer (tensor).} Expression at this layer encompasses the different activation of attention heads and latent states across the model's architecture, contributing to diverse outputs. 
\end{itemize}

For the expression type $e$, the expression distribution produced by $\mathcal{M}$ in response to $\boldsymbol{x}$ can be defined as follows:

\begin{equation}
    O_e(\mathcal{M}, \boldsymbol{x}), \quad e \in \{\text{response}, \text{decoding},\text{latent}\}
    \label{eq:observation_distribution}
\end{equation}

By sampling from this distribution, we can obtain a sampling set with potentially repeated elements:

\begin{equation}
    \mathcal{Y}= \{ \boldsymbol{y}_1, \boldsymbol{y}_2, \ldots, \boldsymbol{y}_n \}, \quad \boldsymbol{y}_i \sim O_e(\mathcal{M}, \boldsymbol{x})
\end{equation}

Here, $\boldsymbol{y}_i$ represents the $i$-th sample obtained from $O_e(\mathcal{M}, \boldsymbol{x})$. With this sampling set, various methods can be employed to estimate the consistency of these expressions. For example, as shown in Fig.~\ref{fig:inconsistent_responses}, we can obtain $\mathcal{Y}= \{ 4,3,3,3,4 \}$. Below are two relatively trivial estimation methods. From a statistical perspective, we can compute the negative variance as a measure of consistency, as shown in Eq.~\ref{eq:dev}; from an information-theoretic perspective, we can use the negative entropy as a measure of consistency, as shown in Eq.~\ref{eq:info}. However, simple variance and entropy may not provide useful guidance for better result updates, and their applicability is limited to tasks where expressions are numerical labels.

\begin{equation}
    -D(\mathcal{Y})=-E(\mathcal{Y}-E(\mathcal{Y}))^2=-0.24
    \label{eq:dev}
\end{equation}
\begin{equation}
    -H(\mathcal{Y}) = \sum_{i=1}^n p(\boldsymbol{y}_i) \log_2 p(\boldsymbol{y}_i) \approx -0.971
    \label{eq:info}
\end{equation}

We will comprehensively discuss existing methods for acquiring consistency signals in Section~\ref{sec:consistency_signal_acquisition}. Those methods may be more helpful.

\begin{figure}[t!]
    \centering
    \includegraphics[width=\linewidth]{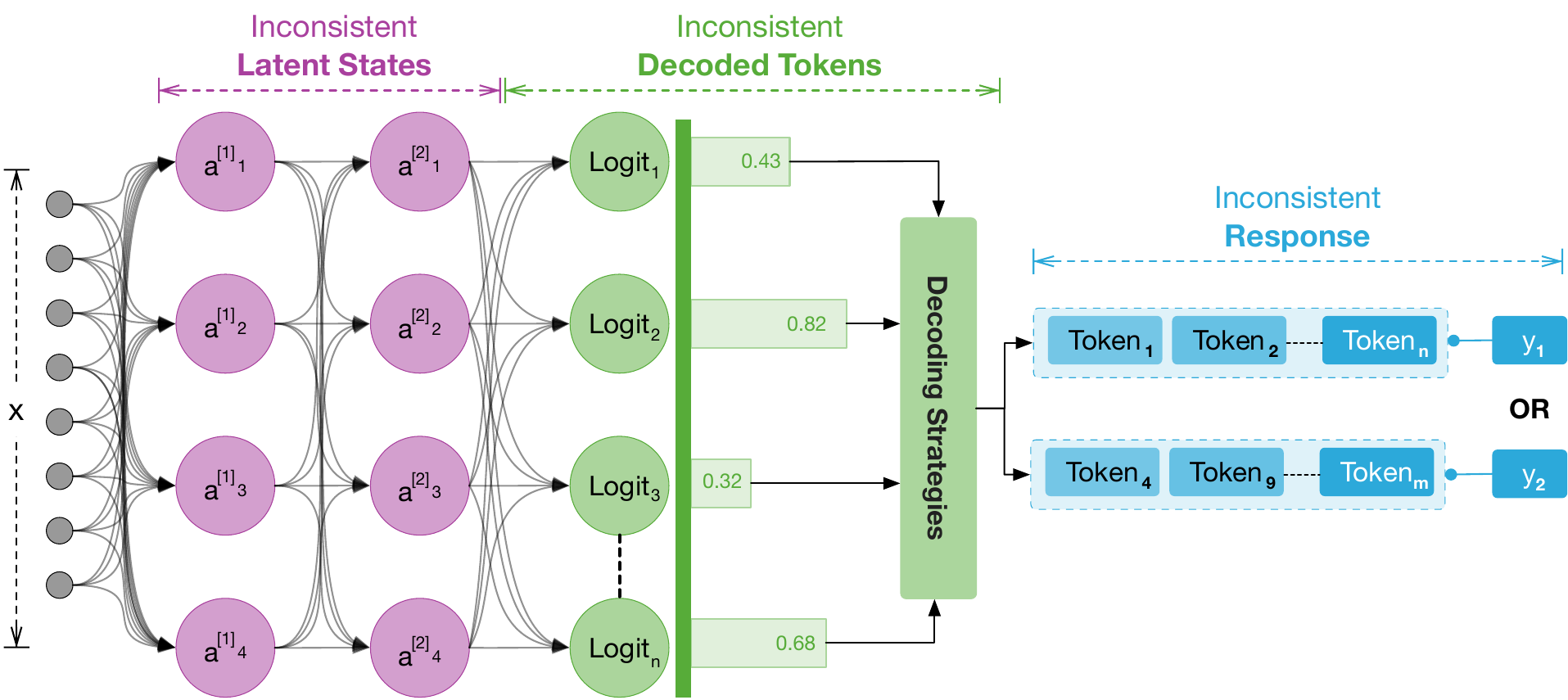}
    \caption{Positions of the Three Types of Consistency}
    \label{fig:consistency_type}
\end{figure}

Additionally, the three different types of ``expressions'' mentioned above constitute the main focus of this paper's discussion on three types of consistency: Response Consistency, Decoding Consistency, and Latent Consistency. Fig.~\ref{fig:consistency_type} visually illustrates the positions of these three types in an LLM.

\subsection{The Hourglass Evolution of Internal Consistency}  \label{sec:hourglass}

\noindent In this section, we delve deeper into the three different types of internal consistency. We conducted a simple experiment where Llama3-8B-Instruct\footnote{\url{https://ai.meta.com/blog/meta-llama-3/}} was asked to respond to a straightforward query many times to observe the consistency of different types of expressions in the $\{\text{response}, \text{decoding}, \text{latent}\}$ layers. And, the given query is: How many full stops (periods) are there: ``.!..!..!''. Below are the methods for collecting sampling sets at different layers. Refer to our \href{https://github.com/IAAR-Shanghai/ICSFSurvey}{GitHub repository} for detailed experimental settings and results.

\textbf{Response Layer.} We used Top-p sampling with a fixed temperature to sample five times. To induce diverse responses, CoT prompting was enabled. We observed the model's final textual choices during free generation. One example output is: ``Let's think step by step. There is one period at the end of the first part, ... So, there are 3 periods in total.'' The resulting sampling set is $\mathcal{Y}_\text{response} = \{ 5,3,3,3,3 \}$.

\textbf{Decoding Layer.} We used five decoding strategies to sample and observe the tokens selected. These decoding strategies included Greedy Decoding, Beam Search Decoding, Sampling Decoding, Top-k Sampling Decoding, and Top-p Sampling Decoding. The sampling set is $\mathcal{Y}_\text{decoding} = \{ 4,4,3,4,4 \}$.

\textbf{Latent Layer.} We hypothesized that different attention heads lead to different answers. To test this, we kept only the $h$-th attention head of the $l$-th Transformer block of model $\mathcal{M}$ active and set the attention output of other heads in that layer to zero, observing which token had the highest probability in the forward pass. We used six different combinations of $l$ and $n$, i.e., $(l,n) \in \{0, 15, 30\} \times \{0, 16\}$. The resulting ordered sampling set is $\mathcal{Y}_\text{latent} = < 0,0,5,4,4,4 >$\footnote{In this set, smaller $l$ are in front; for the same $l$, smaller $n$ are in front.}.

\begin{figure}[t!]
    \centering
    \includegraphics[width=\linewidth]{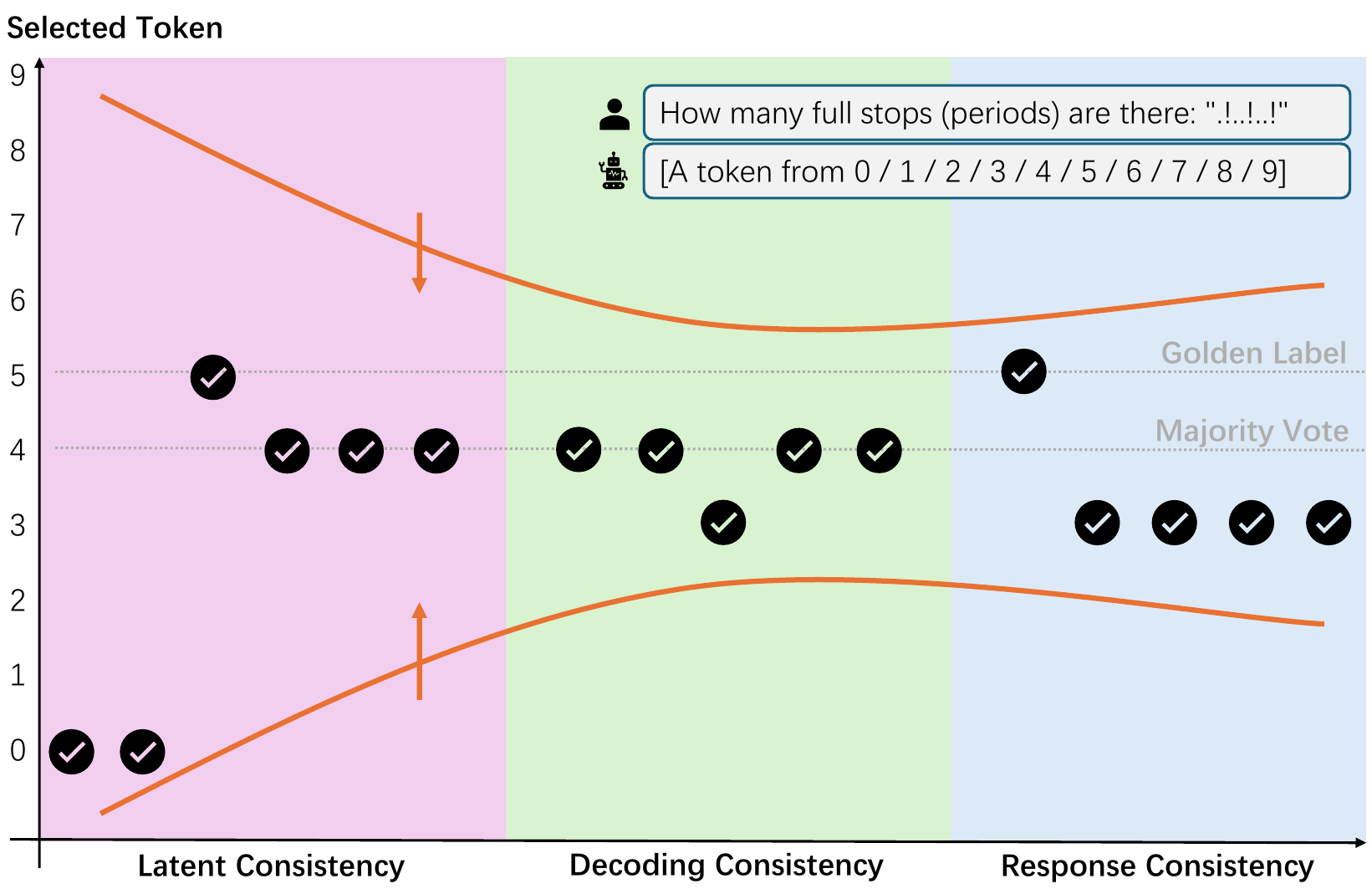}
    \caption{The Hourglass Evolution of Internal Consistency}
    \label{fig:expt}
\end{figure}

The experimental results are also shown in Fig.~\ref{fig:expt}. We observed that the model's answer consistency follows an \textbf{``hourglass evolution'' pattern}, starting from the lower to higher layers at the latent level, passing through the intermediate decoding level, and finally reaching the response level.

We analyze this phenomenon as follows. In the latent state, since the forward propagation is not yet complete, the attention heads near the bottom layers may tend to choose answers randomly. In contrast, the attention heads near the top layers can continually accumulate knowledge due to residual connections, leading to a gradual convergence in judgment. During the decoding phase, all decoding strategies tend to select the token with the higher probability, thus maintaining high certainty. However, at the response stage, greater variability appears. When the LLM generates the first token, it has already conducted reasoning (namely, the latent reasoning~\cite{TheoryLatentReason_24_arXiv_Google}) and made an initial judgment of the answer. However, during the response phase, the output tokens such as ``I'm willing to help.'' can interfere with the model's initial reasoning and preliminary judgment, leading to a collapse of latent reasoning.

From this figure, we can also see that our goal is to have the orange consistency boundary line move as close to the center as possible, which is the goal of internal consistency mining.

\subsection{Status Quo of LLM Consistency} \label{sec:status_quo}

\noindent As indicated at the beginning of the survey, GPT-4o's various responses to the same question (see Fig.~\ref{fig:inconsistent_responses}) already demonstrate that even relatively powerful LLMs still exhibit low consistency. This section examines the current state of LLM consistency from two perspectives.

\textbf{LLMs often provide inconsistent responses, even when they know the correct answer.} The well-known Self-Consistency~\cite{SelfConsistency_23_ICLR_Google} explores the use of the majority voting strategy, where the LLM generates multiple responses and selects the most voted one as the final response. Their experiments showed that on the reasoning benchmark GSM8K~\cite{cobbe2021training}, this method increased the answer accuracy by about 17.9\%. This implies that many initial responses do not represent a consistent answer. In terms of hallucination alleviation, M\"undle et al.~\cite{HalluSelfContradictory_24_ICLR_ETH} proposed the Self-Contradict strategy, which attempts to generate different samples to identify self-contradictory content and then eliminate these contradictions to reduce hallucinations. Their experiment showed that even GPT-4 was able to induce self-contradictions at rates of 15.7\%.

\textbf{LLMs are inconsistent in expressing what they know and do not know, i.e., they lack Self-Knowledge.} For example, Yin et al.~\cite{TheoryKnowUnknown_23_ACL_Fudan} and Cheng et al.~\cite{TheoryKnowUnknown_24_arxiv_Fudan} created datasets consisting of questions that models cannot answer to test whether the models can refuse to answer these questions. Their research showed that models exhibit low consistency in refusing ``I Don't Know'' (IDK) questions, with room for improvement compared to humans.

Therefore, we believe the consistency of results obtained from LLMs using trivial forward propagation, trivial decoding strategies, and trivial model response strategies is low.

% II.E节的图，排版需要放在这里
\begin{figure*}[t!]
    \centering
    \includegraphics[width=\linewidth]{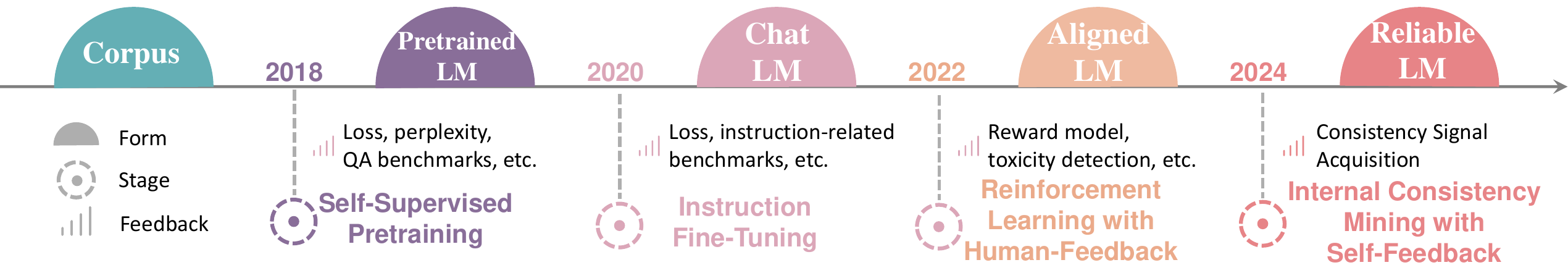}
    \caption{Various Alignments Involved in the LLM development}
    \label{fig:full_consistency}
\end{figure*}

\subsection{Sources of Low Internal Consistency}  \label{sec:sources_of_low}

\noindent Why models exhibit low internal consistency. Here we present some relevant explorations. Understanding these causes can help researchers better improve model performance.

\textbf{Great sensitivity to specific prompts.} Xie et al.~\cite{CalibIC_24_arXiv_SJTU} found that different CoT prompts led to significant differences in latent state distances between intermediate and final layers, affecting consistency. Liu et al.~\cite{LiTM_24_TACL_Stanford} observed a ``lost-in-the-middle'' phenomenon, where models inconsistently respond to prompts based on the position of answers within the long context. Liu et al.~\cite{FFLM_23_NIPS_CMU} further analyzed hallucinations within long contexts. They analyzed that this is caused by the soft attention mechanism, where attention weights become overly dispersed as sequence length increases, leading to poor consistency in reasoning paths.

\textbf{Deficiencies of reasoning.} Yang et al.~\cite{TheoryLatentReason_24_arXiv_Google} investigated whether models use intermediate latent reasoning for answering questions and if strengthening this reasoning could boost accuracy. Their findings revealed that while models do have latent reasoning abilities, these are weak. Enhancing the signal strength of intermediate entities did not significantly improve the model's responses, suggesting current LLM architectures struggle with latent reasoning and may make near-random predictions due to insufficient latent reasoning. Additionally, Zhang et al.~\cite{TheorySnowball_23_arXiv_NYU} argued that models could hallucinate due to the ``snowball effect''. The full attention mechanism makes LLMs overly confident in their outputs, leading to compounding errors if an initial reasoning mistake occurs. Consequently, model's responses may become inconsistent with the knowledge it has learned.

\textbf{Theoretical hypotheses.} Bender et al.~\cite{Parrots_21_FAccT_UoW} proposed that LLMs might be ``stochastic parrots'', learning rules and patterns from training data rather than truly understanding the grammar and semantics of natural language. This inherent randomness in generation reflects a form of internal inconsistency in the model. Ma et al.~\cite{PrincipleSC_22_FITEE} proposed the Principle of Self-Consistency for intelligent agents, aiming to find a coherent model that minimizes internal differences between observed and regenerated data. They found many factors that could affect internal consistency, such as mode collapse\footnote{Mode collapse: A generative model starts producing very similar or repetitive outputs during training, failing to capture the diversity of the data.}, neural collapse\footnote{Neural collapse: The model learns the simplest representation to map input to output, without capturing the complex logic within the data.}, and over-fitting or under-fitting caused by overly high or low dimensional feature spaces.

% In conclusion, both theoretical and experimental findings indicate that model architecture, training processes, and user queries can all contribute to low internal consistency. 

% 未来可能可加的：
% 不可自我认识：无法检查自己的logit
% 幻觉是人类语言的本质，我们依赖幻觉

\subsection{How to Understand Internal Consistency?}

\noindent If there is internal consistency, there must also be corresponding external consistency as illustrated in Fig.~\ref{fig:full_consistency}. Each stage of alignment plays a unique role. Among these alignments, internal consistency is crucial for AI reliability~\cite{han2024llm, TheoryKnowKnow_22_arXiv_Anthropic}:

\begin{itemize}
    \item \textbf{Truthfulness}. LLMs provide factually accurate information, including finding, using, and evaluating source materials correctly.
    \item \textbf{Calibration}. LLMs' probabilistic predictions correspond with frequencies of occurrence.
    \item \textbf{Self-Knowledge}. LLMs know what they know and make accurate predictions about their own behavior.
    \item \textbf{Explainability}. LLMs reveal their ``thinking'' completely and faithfully.
    \item \textbf{Non-deceptiveness}. LLMs are ensured not to lie, even when human preference encourages systematic mistakes or provides rewards for pleasant misconceptions.
\end{itemize}

\section{Self-Feedback Framework} \label{sec:self_feedback}

\subsection{Formulation}

\noindent Self-Feedback is a theoretical framework we have summarized from numerous studies. It includes Self-Evaluation and Self-Update, as shown in the middle part of Fig.~\ref{fig:article_framework}.

\begin{tcolorbox}[colback=white!98!black,colframe=white!30!black,boxsep=1.1pt,top=6pt]
\textbf{Self-Feedback}\\[-0.575em]
\noindent\makebox[\textwidth]{\rule{\textwidth}{0.4pt}}
\\[0.25em]
Narrowly speaking, Self-Feedback refers to the method of improving a model's own internal consistency through its feedback, where ``own'' refers to a specific model instance or a specific response. \\
\\
Broadly speaking, ``own'' can be extended to other models. For example, multiple different models can improve their capabilities through feedback generated from debates among them, which is a more generalized interpretation of Self-Feedback.
\end{tcolorbox}

Based on the above descriptive definition, we can formalize the process of Self-Feedback. For a given model $\mathcal{M}$, query $\boldsymbol{x}$, and a sampling set $\mathcal{Y}$ obtained under a certain expression type, Self-Evaluate\footnote{A small number of methods use other models $\text{SelfEvaluate}_\mathcal{N}(\mathcal{Y})$ or even external tools $\text{SelfEvaluate}_\text{tool}(\mathcal{Y})$ during Self-Evaluate.} is first performed to obtain feedback $f$:

\begin{equation}
    f = \text{SelfEvaluate}_\mathcal{M}(\mathcal{Y})
    \label{eq:selfevaluate}
\end{equation}

We can use the obtained feedback $f$ to let the model $\mathcal{M}$ directly update the original expression $\mathcal{Y}$ to $\boldsymbol{y}'$:

\begin{equation}
    \boldsymbol{y}'=\text{SelfUpdate}_\mathcal{M}(\mathcal{Y}, f)
    \label{eq:selfupdate1}
\end{equation}

We can also use the obtained feedback $f$ to select better responses and optimize the model parameters $\mathcal{M}$ through fine-tuning or other strategies to obtain a better model $\mathcal{M}'$:

\begin{equation}
    \mathcal{M}'=\text{SelfUpdate}_\mathcal{M}(\mathcal{Y}, f)
    \label{eq:selfupdate2}
\end{equation}

Additionally, we can use the feedback to update other models, such as updating a student model $\mathcal{N}$:

\begin{equation}
    \mathcal{N}'=\text{SelfUpdate}_\mathcal{N}(\mathcal{Y}, f)
    \label{eq:selfupdate3}
\end{equation}

The combination of Self-Evaluate defined in Eq.~\ref{eq:selfevaluate} and Self-Update defined in Eqs.~\ref{eq:selfupdate1}, \ref{eq:selfupdate2}, and \ref{eq:selfupdate3} constitutes various Self-Feedback methods. During Self-Evaluate, external signals may be used, and during Self-Update, other models may be updated. This interaction with external entities is referred to as generalized Self-Feedback.

\subsection{Taxonomy}

\noindent Self-Feedback centers on $\text{SelfEvaluate}$, $\text{SelfUpdate}$, and the feedback signal $f$. Rather than fragmenting the survey by these elements, we classify the papers we read by tasks and lines of work, as shown in Fig.~\ref{fig:article_framework}. The four key tasks are:

\begin{itemize}
\item \textbf{Section~\ref{sec:consistency_signal_acquisition} (Consistency Signal Acquisition)} summarizes methods for obtaining the feedback signal $f$. We consider this task important because many Self-Feedback methods overlook this dimension. For instance, the feedback signal in Self-Consistency~\cite{SelfConsistency_23_ICLR_Google} should be classified under scalar-based Consistency Estimation methods.

\item \textbf{Section~\ref{sec:reasoning_elevation} (Reasoning Elevation)} is one of the key focuses of this paper. We have discussed the distinctions and connections between reasoning and hallucination in Section~\ref{sec:lack_reason_exhibit_hallu}. To clarify, the primary focus here is on Self-Feedback methods aimed at QA tasks.

\item \textbf{Section~\ref{sec:hallucination_alleviation} (Hallucination Alleviation)} is another critical focus of this paper. Here, we concentrate on Self-Feedback methods targeted at open-ended generation tasks. Note: We also provide Table~\ref{tab:reason_hall_paradigms} to share specific lines of work related to Reasoning Elevation and Hallucination Alleviation.

\item \textbf{Section~\ref{sec:other_tasks} (Others)} briefly covers Self-Feedback methods applied to tasks beyond Reasoning Elevation and Hallucination Alleviation, such as knowledge distillation and embedding generation.
\end{itemize}

\begin{table*}[ht!]
\begin{threeparttable}

\caption{Different Lines of Work in Reasoning Elevation and Hallucination Alleviation}
\label{tab:reason_hall_paradigms}
\scriptsize
\begin{tabular}{p{2.4cm}p{1.3cm}p{2.1cm}p{0.8cm}p{0.5cm}p{1.65cm}p{1.5cm}p{3.85cm}}
\toprule
\textbf{Section: Paradigm}  &  \textbf{Expression}     & \textbf{Signal Type}                    & \textbf{\#LLM}  &\textbf{Train.} & \textbf{Self-Evaluation}                     & \textbf{Self-Update}                   & \textbf{Typical Works}                        \\
\midrule
\ref{sec:reasoning_topologically}: Reasoning Topologically                  & Response, \newline Decoding & Scalar, Textual, \newline Contrastive   & 1      & No         & Majority Voting, Value Function & Best Selection                & Self-Consistency~\cite{SelfConsistency_23_ICLR_Google}, ToT~\cite{ToT_23_NeuIPS_Princeton}, GoT~\cite{GoT_24_AAAI_ETH}, Quiet-STaR~\cite{QuietSTaR_24_arXiv_Stanford}           \\
\midrule
\ref{sec:refining_with_responses}: Refining with Responses                  & Response            & Textual                        & 1 or 2 & Half            & Sampling                            & Best Selection, Model Tuning & Self-Improve~\cite{SelfImprove_23_EMNLP_Illinois}, ConCoRD~\cite{ConCoRD_22_EMNLP_Stanford}, LEMA~\cite{LearnMistake_24_arXiv_MS}, Mistake Tuning~\cite{LearnMistake_24_arXiv_UCSD}          \\
\midrule
\ref{sec:multi_agent}: Multi-Agent Collaboration                             & Response            & Textual, Scalar                & $\geq 2$     & Rare   & Negotiation                         & Answer Aggregation            & FORD~\cite{ModalCollaboration_23_EMNLP_HIT}, MACNet~\cite{MACNet_24_arXiv_THU}, REFINER~\cite{REFINER_24_EACL_EPFL}, Multi-Agent Debate~\cite{Debate_23_arXiv_MIT}                \\
\midrule
\midrule
\ref{sec:refining_the_response_iteratively}: Refining the Response Iteratively        & Response            & Textual, External              & 1      & Few           & Model Generate Critique             & Model Generate Refinement     & Self-Refine~\cite{SelfRefine_23_NeuIPS_CMU}, Reflexion~\cite{Reflexion_23_NeuIPS_Northeastern}, Self-Correct~\cite{SelfCorrect_23_ICLR_AI2}, Self-Debug~\cite{SelfDebug_24_ICLR_Google} \\
\midrule
\ref{sec:mitigating_hallucination_while}: Mitigating Hallu. while Generating & Response            & Textual, Contrastive, \newline External & 1      & Few            & Inherent model evaluation           & Model Delete Hallucination    & Self-Contradict~\cite{HalluSelfContradictory_24_ICLR_ETH}, EVER~\cite{EVER_arXiv_23_UNC}, FEVA~\cite{FAVA_24_arXiv_Washington}       \\
\midrule
\ref{sec:decoding_truthfully}: Decoding Truthfully                    & Decoding            & Contrastive                    & 1 or 2 & No            & Evaluate Decoding Path               & Select the Best Decoding Path & DoLa~\cite{DoLa_24_ICLR_MIT}, CAD~\cite{ContextAwareD_23_arXiv_Washington}, DIVER~\cite{DIVER_24_arXiv_IA}, SED~\cite{SED_24_arXiv_FDU}                      \\
\midrule
\ref{sec:activate_truth}: Activating Truthfulness                 & Latent              & Contrastive                    & 1      & No            & Evaluate Latent States              & Activate the Best States      & ITI~\cite{ITI_23_NeuIPS_Harvard}, TrFr~\cite{TrFr_24_AAAI_BUAA}, TruthX~\cite{TruthX_24_ACL_ICT}               \\
\bottomrule
\end{tabular}

\begin{tablenotes}
\footnotesize
\item \textit{Note}: This table summarizes the characteristics of representative methods. The first three lines are dedicated to ``Reasoning Elevation'', while the latter four lines are focused on ``Hallucination Alleviation.'' \#LLM indicates the number of LLMs needed. Train. denotes ``How many works need training?''
\end{tablenotes}

\end{threeparttable}
\end{table*}
% TODO：删除掉了Inference cost，未来可以考虑使用inference-time time complexity和inference-time space complexity来代替。或者单独添加一个推理次数，表示生成多少次响应。

\section{Task: Consistency Signal Acquisition} \label{sec:consistency_signal_acquisition}

% 概念
\noindent Consistency signal acquisition refers to evaluating the consistency of expressions after obtaining the sampling set $\mathcal{Y}$. The evaluated signal can help the model update its expressions or parameters, thereby improving the model's internal consistency. Therefore, it is a pivotal task within the Self-Feedback framework. These methods either require access only to the model's output contents, to the logits, or to the latent states of the model. Depending on the depth of access required by different methods, the approaches mentioned in this section are categorized as black-box (accessing only the model's output contents), gray-box (also accessing logits), and white-box (also accessing the model's latent states). Numerous explorations have been undertaken in this task. These include:

\begin{itemize}
    \item Section~\ref{sec:uncertainty}: Uncertainty Estimation (Scalar)
    \item Section~\ref{sec:confidence}: Confidence Estimation (Scalar)
    \item Section~\ref{sec:hallucination}: Hallucination Detection (Scalar)
    \item Section~\ref{sec:critiquing}: Verbal Critiquing (Textual)
    \item Section~\ref{sec:contrastive_optimization}: Contrastive Optimization (Contrastive)
    \item Section~\ref{sec:external_feedback}: External Feedback (External)
\end{itemize}
% TODO 高优先级：我们把这些工作线也总结在了表xxx中。字段包括：透明性(黑白灰盒)，#llm，reference-free，计算公式，计算目标(uncertainty，confidence，hallucination等)，基于sample与否等

% \textbf{uncertainty estimation} derived from the traditional machine learning era, modern model response \textbf{confidence estimation}, finer-grained \textbf{hallucination detection}, and non-scalar but textual \textbf{verbal critiquing}. There are also broader or relatively implicit types such as \textbf{contrastive optimization} and \textbf{external feedback}.

% 前三条工作线的联系与区别
The first three lines are actually quite similar. They all provide scalar feedback for LLM responses, and some works even mix the keywords from these three lines, such as \cite{Uncertainty_21_EACL_UCSB, Uncertainty_24_TMLR_Illinois, BelieveOrNot_24_arXiv_Google}. The main difference lies in their downstream tasks. Estimating uncertainty and confidence are two sides of the same coin, both assessing the model's certainty on a $[0,1]$ scale to optimize reasoning. While hallucination detection identifies hallucinations from $\{0, 1\}$, primarily aimed at alleviating hallucinations.

% 后三条工作线
In addition to the aforementioned works that obtain scalar signals, other types of signals have been explored. Verbal Critiquing refers to having the language model directly evaluate the quality of an output, providing suggestions for improvement. External Feedback leverages external sources, such as textual feedback from other robust models or error messages from a compiler in code generation tasks. Finally, there is a more implicit signal, contrastive optimization, which obtains consistency signals through the comparison between different expressions and optimizes towards consistency.

% 讲解范围
In this section, we focus more on the first three lines of work, as they are often studied independently and are hotspots in academic research. The last three lines of work are only briefly mentioned here, as they tend to be relatively simple or implicit methods. They will be elaborated in Sections~\ref{sec:reasoning_elevation},~\ref{sec:hallucination_alleviation}.

\subsection{Uncertainty Estimation} \label{sec:uncertainty}

% 定义
\noindent Uncertainty estimation refers to estimating the data uncertainty, model uncertainty, and distributional uncertainty involved in the neural networks~\cite{deng2023uncertainty}.

% 不确定性的建模/来源
For uncertainty estimation in the NLP field, Hu et al.~\cite{SurveyUncertainty_23_arXiv_Nankai} conducted a detailed survey. They categorize sources and modeling methods of uncertainty into three approaches:  1) \textbf{Calibration Confidence-based Methods}: This approach compares the accuracy of predicted probabilities with actual probabilities. 2) \textbf{Sampling-based Methods}: This approach models the variability of multiple expressions provided by the model, allowing us to observe the arising uncertainties. This method is also the focus of our article. 3) \textbf{Distribution-based Methods}: This approach evaluates inherent uncertainty by analyzing the dataset's distribution characteristics.

% MCD方法
We introduce an important method cluster within Sampling-based Methods: Monte Carlo Dropout (MCD)~\cite{pmlr-v48-gal16}. General deep neural network predictions are often deterministic, and multiple samples yield consistent answers, preventing us from understanding the model's implicit certainty about the results. The MCD method uses the dropout technique to construct an implicit binomial distribution. For example, a 50\% dropout probability constructs a $ B(\text{\#activation}, 0.5)$ binomial distribution, which implicitly creates multiple models with different parameters $\theta_i \sim q(\theta), i=1,2,\ldots,n$. At test time, MCD uses multiple models to obtain multiple output results $P(\boldsymbol{y}_i | \boldsymbol{x}; \theta_i)$ and estimates the uncertainty by calculating the variance of results. As for LLM, obtaining different expressions is much easier, such as using temperature coefficients. From the perspective of MCD, changing the temperature (values of the Softmax layer) implicitly constructs different models.

% 简单的方法
Besides MCD, which offers more explanatory insights, there are simpler, Sampling-based Methods available. For example, the Active Prompting strategy proposed by~\cite{ActivePrompt_23_arXiv_HUST} uses disagreement in answers as an estimate of uncertainty, $\text{SelfEvaluate}(\mathcal{Y}) \triangleq \frac{|\text{unique}(\mathcal{Y})|}{|\mathcal{Y}|}$. Here, $\text{unique}(\mathcal{Y})$ represents the set after removing duplicate elements.

% TODO 不重要目前 如果要加更多文献，uncertainty综述里，还有active prompting文章里还有很多内容可选

\subsection{Confidence Estimation} \label{sec:confidence}

% 引入
\noindent Confidence is the opposite of uncertainty, focusing on reliability scores to enhance user trust.

% 重点方法阐述
In this line of work, Self-Evaluation is the core method\footnote{The Self-Evaluation~\cite{TheoryKnowKnow_22_arXiv_Anthropic} here denotes a method, not the Self-Evaluation module in Self-Feedback framework. To distinguish between the two, a citation marker will be appended when referring to the method.}. The concept of Self-Evaluation was first proposed in~\cite{TheoryKnowKnow_22_arXiv_Anthropic}, where the goal is for the model to express its level of confidence using its own knowledge and reasoning. As shown in Fig.~\ref{fig:prompt_for_self_eval}, the Self-Evaluation method simply asks the model: Is the proposed answer True or False? Then, the confidence score, P(True), is extracted from the model's logits.

\begin{figure}[h!]
\centering

\begin{tcolorbox}[colback=blue!5!white,colframe=blue!75!black,fontupper=\footnotesize,fonttitle=\scriptsize]
\textbf{Question}: Who was the first president of the United States? \\
\textbf{Proposed Answer}: George Washington was the first president. \\
\textbf{Is the proposed answer}: \\
(A) True \\
(B) False \\
\textbf{The proposed answer is}:
\end{tcolorbox}

\caption{Prompt for Self-Evaluation~\cite{TheoryKnowKnow_22_arXiv_Anthropic}}
\label{fig:prompt_for_self_eval}
\end{figure}

Besides naively asking the model whether it thinks the proposed answer is correct, some works have proposed other frameworks. For instance, BSDetector~\cite{TheoryUncertainty_23_arXiv_UoMaryland} is a confidence estimation framework suitable for both black-box and white-box models. It combines the consistency of multiple outputs sampled from the model with the model's own reflection on its output, weighting these scores to obtain the confidence scores. Another example, TrustScore~\cite{TrustScore_24_arXiv_UoEdinburgh} is a reference-free confidence estimation framework using behavior consistency. It generates distractors based on entity information rules from Wikipedia, asks the LLM multiple times, and checks if it consistently chooses its own generated answer.

\subsection{Hallucination Detection} \label{sec:hallucination}

\noindent Hallucination Detection aims to identify untruthful or unfaithful text within a response.  SelfCheckGPT~\cite{HalluSelfCheckGPT_23_EMNLP_Cambridge} provides a reference-free hallucination detection framework. Specifically, the goal of SelfCheckGPT is to determine the presence of hallucination in a given query $\boldsymbol{x}$ and response $\boldsymbol{y}_0$. The framework works in three steps. Firstly, the model samples several different responses, $\mathcal{Y}= \{\boldsymbol{y}_1, \boldsymbol{y}_2, \ldots, \boldsymbol{y}_n \}$. Secondly, it calculates whether $\boldsymbol{y}_{1:n}$ support $\boldsymbol{y}_0$. Finally, it summarizes the support level to calculate the final score. Designing support level metric is where creativity can be applied, and the authors provide five different methods:

\begin{itemize}
    \item \textbf{Similarity-based}: Compute the negation of the mean similarity between $\boldsymbol{y}_{1:n}$ and $\boldsymbol{y}_0$;
    \item \textbf{QA-based}: Generate many questions from $\boldsymbol{y}_0$ and test consistencies in the answers derived from $\boldsymbol{y}_0$ and $\boldsymbol{y}_{1:n}$;
    \item \textbf{N-gram model-based}: Build an n-gram model from $\mathcal{Y}$, then use it to compute the negation of the mean transition probability between tokens in $\boldsymbol{y}_0$.
    \item \textbf{Natural language inference (NLI)-based}: Compute the mean probability of contradiction between the responses;
    \item \textbf{Prompt-based}: Similar to Self-Evaluation~\cite{TheoryKnowKnow_22_arXiv_Anthropic}, directly ask the language model whether $\boldsymbol{y}_{1:n}$ support $\boldsymbol{y}_0$.
\end{itemize}

Beyond the extensive methods of SelfCheckGPT, there are other interesting approaches as well. The Alibaba team proposed INSIDE~\cite{INSIDE_24_ICLR_Alibaba} for deeper exploration. They sampled latent vectors from the intermediate layers and calculated the covariance matrix of these vectors. Since the eigenvalue of the covariance matrix represents data variability, they used this value as a measure of hallucination. Additionally, some methods utilize multiple agents to detect hallucinations. For example, Cross Examination~\cite{CrossExamine_23_EMNLP_TAU} employs two LLMs, an Examinee and an Examiner, using a cross-examination approach to determine factual errors.

\subsection{Verbal Critiquing}  \label{sec:critiquing}

\noindent Inspired by the idea that ``all tasks are generation tasks''~\cite{BARTScore_21_NeuIPS_CMU,GPTScore_23_arXiv_NUS}, many works have proposed allowing LLMs to \textit{generate} more semantically rich textual signals. These include:

\textbf{Let LLMs offer critiques.} Saunders et al.~\cite{SelfCritiquing_22_arXiv_OpenAI} use a fine-tuned Self-Critiquing model to generate insights on content. McAleese et al.~\cite{CriticGPT_24_arXiv_OpenAI} use RLHF based on the GPT-4 model to train the model to critique code generation, resulting in CriticGPT. Du et al.~\cite{Debate_23_arXiv_MIT} propose the Multi-Agent Debate method, where two agents generate modifications to each other's content, gradually converging to an outcome.

\textbf{Let LLMs summarize.} Xiong et al.~\cite{ModalCollaboration_23_EMNLP_HIT} use a Judge LLM to aggregate the results produced by multiple agents, providing a final judgment. Graph-of-Thought~\cite{GoT_24_AAAI_ETH} uses the aggregation of thoughts to perform subsequent reasoning.

\textbf{Let LLMs refine the text.} These methods involve the LLM generating a refined response as a better result~\cite{SelfRefine_23_NeuIPS_CMU,Reflexion_23_NeuIPS_Northeastern,SelfCorrect_23_ICLR_AI2}.

\subsection{Contrastive Optimization}  \label{sec:contrastive_optimization}

\noindent Contrastive optimization is an implicit signal acquisition method, which often involves constructing a scoring function, $\text{score}(\boldsymbol{y}_i)$, to evaluate all responses in the sampling set $\mathcal{Y}$, $\{ \text{score}(\boldsymbol{y}_i) | i=1,2,\ldots,n\}$. Finally, the best candidate is selected as $\boldsymbol{y}_\text{best} = \argmax_{\boldsymbol{y}_i}{\text{score}(\boldsymbol{y}_i})$.

\textbf{At the latent layer}, in order to find attention heads with a stronger preference for truthfulness, Li et al.~\cite{ITI_23_NeuIPS_Harvard} trained a probe to evaluate the attention head's ability to answer questions truthfully. \textbf{At the decoding layer}, Self-Evaluation~\cite{TheoryKnowKnow_22_arXiv_Anthropic} can be used to evaluate the reasoning paths during beam search, comparing scores to choose a better decoding direction~\cite{SelfEvaluation_23_NeuIPS_NUS}. \textbf{At the response layer}, Self-Consistency~\cite{SelfConsistency_23_ICLR_Google} strategy implicitly relies on comparisons between different responses. A variant, Soft Self-Consistency~\cite{SoftSelfConsistency_24_arXiv_UNCChapel}, calculates the joint probability of tokens for each response as the scoring function.

\subsection{External Feedback}  \label{sec:external_feedback}

\noindent Sometimes, feedback from the model itself is not sufficient. For example, in code generation, if there are hallucinations (bugs) in the code, it is difficult for even humans to accurately identify some bugs without executing the code with an external executor. Self-Debug~\cite{SelfDebug_24_ICLR_Google} proposes using the execution results from an external executor as feedback. Besides using external tools, some works use other models as external feedback sources, such as a more powerful teacher model~\cite{PERsD_23_EMNLP_NTU} or a peer model~\cite{Debate_23_arXiv_MIT}. The commonly used RAG method, which can incorporate information retrieved from external sources as external feedback, is another example utilizing external feedback. % However, as mentioned in Section~\ref{sec:out_of_scope}, using external information sources is not the focus of this paper as it does not pertain to Internal Consistency Mining. Nevertheless, many studies employing external signals in Self-Feedback are summarized in Section~\ref{sec:other_tasks}.

\section{Task: Reasoning Elevation} \label{sec:reasoning_elevation}

\noindent Reasoning Elevation refers to enhancing the logical reasoning capabilities of language models during response generation to improve their internal consistency. The primary feature of this line of work is the use of benchmarks in the form of QA tasks. We have identified three significant lines of work, as shown in the upper part of Table~\ref{tab:reason_hall_paradigms}.

\subsection{Reasoning Topologically} \label{sec:reasoning_topologically}

\begin{figure*}[t!]
    \centering
    \includegraphics[width=\linewidth]{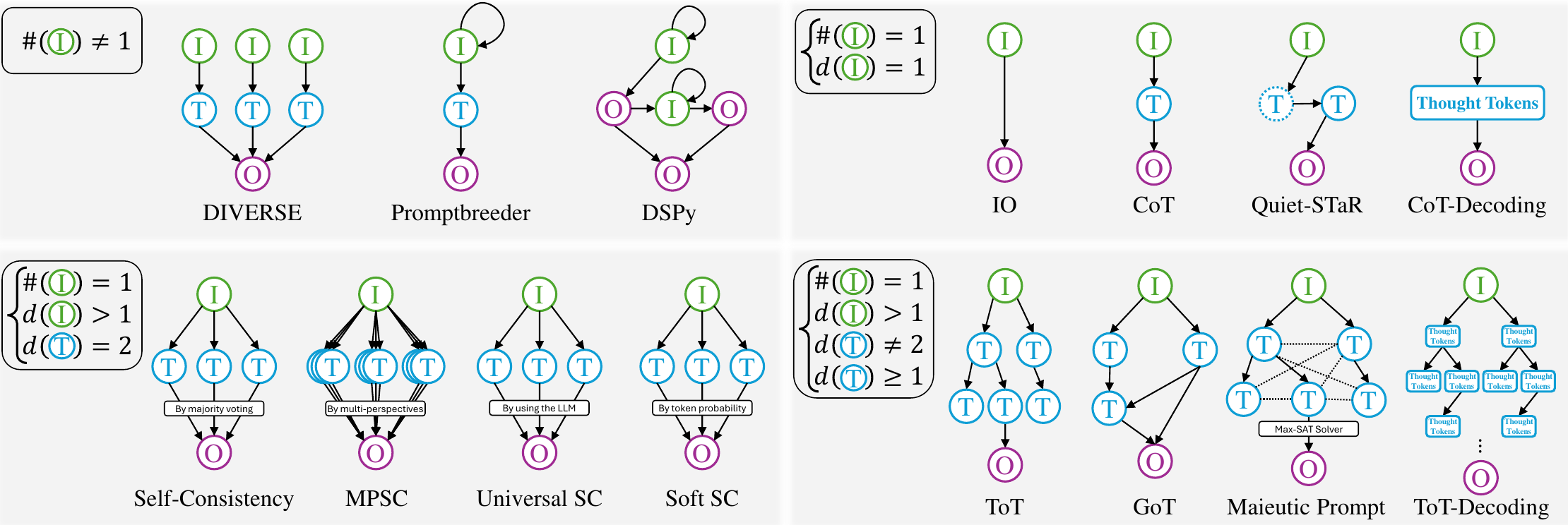}
    \caption{Different Reasoning Topologies. \circledColoredChar{darkgreen}{I}/ \circledColoredChar{darkblue}{T}/ \circledColoredChar{darkpurple}{O} indicate input / intermediate thought / output, respectively. $\#(\cdot)$ and $d(\cdot)$ indicate the number and the degree of nodes, respectively.}
    \label{fig:reason_topo}
\end{figure*}

% 引入
\noindent When answering a question, LLMs may choose different reasoning paths, but not all reasoning paths lead to the correct answer. Therefore, finding reasoning paths that are consistent with the learned knowledge becomes a key issue, leading to a series of works focusing on optimizing reasoning paths. Fig.~\ref{fig:reason_topo} summarizes the similarities and differences of these works.

% X-of-Thought 方法简介
A survey~\cite{SurveyXofThought_24_arXiv_ETH} covers various X-of-Thought (XoT) methods. \textbf{Input-Output (IO)} is the simplest approach, asking a question and getting an answer directly, but often struggles with complex problems. To address this, \textbf{Chain-of-Thought (CoT)}~\cite{RealCoT_22_NeuIPS_Google} was introduced, adding intermediate reasoning steps, though errors in reasoning can affect results. \textbf{Self-Consistency (SC)}~\cite{SelfConsistency_23_ICLR_Google} improves accuracy via majority voting but is limited in exploratory power. \textbf{Tree-of-Thought (ToT)}~\cite{ToT_23_NeuIPS_Princeton} views reasoning as a path with multiple successor nodes for deeper exploration, while \textbf{Graph-of-Thought (GoT)}~\cite{GoT_24_AAAI_ETH} aggregates reasoning chains across nodes. Similar to GoT, \textbf{Maieutic Prompting}~\cite{Maieutic_22_EMNLP_UoWashington} builds entailment relationships between thoughts, then constructs a Max-SAT~\cite{Battiti2009} problem to obtain the best choices.

% 解决问题：只能聚合具有固定Label集合的query
Most XoT methods require sampling and aggregation of thoughts, often limited to queries with fixed label sets during aggregation. To solve this problem, several works have emerged. \textbf{Multi-Perspective Self-Consistency (MPSC)}~\cite{MPSC_23_arXiv_PKU} targets code generation tasks, evaluating each solution from multiple perspectives (solution, specification, and test case) to select the best one. \textbf{Universal Self-Consistency (Universal SC)}~\cite{UniversalSelfConsistency_23_arXiv_Google} uses LLMs instead of simple answer matching to choose the most selected response, enhancing the stability of the majority voting. \textbf{Soft Self-Consistency (Soft SC)}~\cite{SoftSelfConsistency_24_arXiv_UNCChapel} proposes a more adaptive scoring function, calculating the joint probability of tokens in a response as the scoring function, thus extending the problem scope to soft labels.

% 解决问题：隐含推理和显式推理的矛盾
Additionally, \textbf{Quiet Self-Taught Reasoner (Quiet-STaR)}~\cite{QuietSTaR_24_arXiv_Stanford} addresses the issue mentioned in Section~\ref{sec:hourglass}, where ``although complex reasoning in responses is beneficial for solving intricate problems, they may disrupt model's latent reasoning due to redundant reasoning text, thereby increasing response-level inconsistency.'' Quiet-STaR samples rationales from the model's responses and wraps each rationale between special markers, that is, \textless \textbar startofthought\textbar\textgreater\ and \textless \textbar endofthought\textbar\textgreater, to assist next-token reasoning. These rationales are invisible to the user, making latent reasoning explicit and effectively reducing conflicts.

% 推理拓扑拓展到input阶段
However, these lines of work are mostly focused on how to choose the next thought from an input, overlooking the input stage. An input is a combination of a query and a prompt template. While the query remains relatively unchanged, the instructions and demonstrations in the prompt template can be optimized. Several works have explored this area: \textbf{DIVERSE}~\cite{SelfTeach_23_ACL_PKU} pre-constructs various prompt templates to increase prompt diversity. \textbf{Promptbreeder}~\cite{Promptbreeder_23_arXiv_DeepMind} uses genetic algorithms~\cite{genetic_algorithm} to continuously optimize the original prompt template. \textbf{DSPy}~\cite{DSPy_24_ICLR_Stanford} innovatively builds a prompt optimizer, similar to a gradient optimizer in PyTorch. These methods extend reasoning topology to the input stage, demonstrating significant creativity. Boldly, we could construct a reasoning-topology-oriented framework incorporating prompt optimization, which could potentially solve more complex problems.
% TODO: 可以考虑把这个放到 future directions里

% ssc here

% 推理拓扑拓展到decoding阶段
Furthermore, we can extend our approach to the decoding stage. \textbf{CoT Decoding}~\cite{CoT_24_arXiv_Google} incorporates CoT's ideas into the decoding process, attempting to identify CoT-included decoding paths in the natural decoding process. \textbf{ToT Decoding}~\cite{SelfEvaluation_23_NeuIPS_NUS} integrates ToT concepts into decoding, replacing beam search criteria with Self-Evaluation~\cite{TheoryKnowKnow_22_arXiv_Anthropic}, where each token's selection depends on confidence scores $C(\cdot)$, achieving better reasoning, as shown in Eq.~\ref{eq:beam_with_selfeval}, where $\boldsymbol{y}^t$ is the $t$-th token in string $\boldsymbol{y}$.

\begin{equation}
    P(\boldsymbol{y}) = \prod_t P(\boldsymbol{y}^t | \boldsymbol{y}^{1:t-1})C(\boldsymbol{y}^t)
    \label{eq:beam_with_selfeval}
\end{equation}

% Self-Evaluation策略
\textbf{Self-Evaluation Strategy.} The methods discussed in this section typically require searching the thought graph, necessitating evaluators to determine the usefulness of thoughts and whether they merit further exploration. These works generally use three approaches: Majority Voting, selecting the most consistent response among multiple thoughts~\cite{SelfConsistency_23_ICLR_Google}; Rule-based methods, designing specific scoring functions based on the problem, such as error scoring functions in sorting tasks, representing the number of inversions and frequency differences before and after sorting~\cite{GoT_24_AAAI_ETH}; and LLM-based methods, like the scoring function in the Game of 24 task, where LLMs rate the solution's feasibility as ``sure/maybe/impossible''~\cite{ToT_23_NeuIPS_Princeton}.

% Self-Update策略
\textbf{Self-Update Strategy.} For Self-Consistency prompting, the update uses a majority voting result. For ToT prompting, the update method uses BFS and DFS strategies to search and select suitable thoughts as output. For GoT prompting, the update method is similar to ToT but includes more extensive search spaces, aggregating different thoughts.

% 方法评价
Despite the innovations, these methods have several limitations~\cite{SurveyXofThought_24_arXiv_ETH}: 1) They often select extremely simple tasks like Game of 24, Sorting, and Keyword Counting for experiments. 2) They incur high reasoning costs. 3) They struggle to adapt to general tasks and deployment.

\subsection{Refining with Responses} \label{sec:refining_with_responses}

\noindent Refining with Responses refers to the process where an LLM first generates multiple responses, then identifies the better responses or self-evaluates its own generated content and corrects errors, and finally refines its output or fine-tunes the model itself to improve response consistency. The following are three common lines of work.

\textbf{Fine-tuning from the collected responses.} This line of work involves ``using self-generated data to fine-tune itself.'' Specifically, they often use LLMs to produce multiple answers, select the better responses from them, and then use these better responses to fine-tune the model, enhancing its reasoning capabilities. For example, Self-Improve~\cite{SelfImprove_23_EMNLP_Illinois} uses a majority voting strategy to obtain better outputs, collecting such data to fine-tune the model itself. Similarly, Tian et al.~\cite{SelfImprovement_24_arXiv_Tencent} propose a framework called Self-Improvement, which uses Monte Carlo Tree Search for data synthesis while generating fine-tuning datasets, improving model's reasoning capabilities. % This concept is not only effective in the reasoning domain but also finds applications in other fields like the web agent domain. Self-Improved Agents~\cite{WebAgent_24_arXiv_UPenn} improved performance by 31\% using this method. In the preference optimization field, SRPO (Self-Improving Robust Preference Optimization)~\cite{SelfImprove_24_arXiv_Cohere}, and Self-Alignment~\cite{SelfAlignment_23_NeuIPS_CMU} both utilize model-generated preferences to align with human preferences.

\textbf{Learning from mistakes.} This line of work is similar to fine-tuning from the collected responses but focuses on learning from errors and optimizing by avoiding mistakes. This intuitive method naturally improves model performance by avoiding errors. For instance, the LEMA (LEarning from MistAkes) method proposed by~\cite{LearnMistake_24_arXiv_MS} samples multiple reasoning rationales, has GPT-4 annotate and correct errors among them, and uses the corrected rationales to form a new dataset for re-fine-tuning the model. Similarly, Tong et al.~\cite{LearnMistake_24_arXiv_UCSD} propose the Mistake Tuning scheme: it has the model self-rethink and correct its errors based on references, using large amounts of such self-corrected datasets to fine-tune the model.

\textbf{Getting better response with NLI models.} Besides fine-tuning methods, we also demonstrate rule-based optimization techniques using NLI~\cite{NLIConsistency_22_CS224N_Stanford,ConCoRD_22_EMNLP_Stanford}. With an NLI model, we can identify the relationships between multiple samples and find better responses. For instance, Agarwal et al.~\cite{NLIConsistency_22_CS224N_Stanford} use a pre-trained NLI model to identify and correct logically inconsistent statements generated by a pre-trained language model. They then convert the entailment and contradiction probabilities of the NLI into a Max-SAT problem~\cite{Battiti2009}, and use a constraint solver~\cite{ignatiev2019rc2} to optimize and obtain more accurate and consistent predictions.

\subsection{Multi-Agent Collaboration}  \label{sec:multi_agent}

\noindent The methods in this category generally involve using more than one LLM to collaboratively solve problems, address contradictions, and promote consistency, essentially constituting a generalized form of Self-Feedback. There are numerous papers in the Multi-Agent field; here, we list some typical and novel works that employ Multi-Agent systems for Self-Feedback. For a more comprehensive understanding, refer to the extensive survey on LLM Agents by Wang et al.~\cite{SurveyAgent_24_Frontiers_RUC}.

\textbf{Debate Frameworks.} Multi-Agent Debate~\cite{Debate_23_arXiv_MIT} utilizes multiple peer models that engage in iterative debates, with a fixed number of rounds as the stopping condition. Their experiments show that debates with three or fewer rounds can generally lead to convergence among agents (i.e., LLMs consistently agreeing on the same answer). Xiong et al.~\cite{ModalCollaboration_23_EMNLP_HIT} further propose the FORD (Formal Debate Framework), which introduces a Judge LLM to summarize the agents' statements at the end, also using a fixed number of rounds as the stopping condition. They expand the scope of LLM debates by exploring the effects of debates among models with mismatched capabilities in various scenarios. REFINER~\cite{REFINER_24_EACL_EPFL} trains two models with different roles: a generator for intermediate reasoning steps and a critic for feedback, continuing the iterative dialogue until the correct answer is obtained or the critic has no further feedback. Notably, using the correct answer as a stopping condition has been criticized as unrealistic~\cite{TheoryNoReason_24_ICLR_Google}. 

\textbf{Game-Theoretic Approaches.} The Consensus Game proposed by Jacob et al.~\cite{ConsensusGame_24_ICLR_MIT} deviates from the above frameworks by avoiding direct dialogue between LLMs. Instead, different LLMs participate in a game, based on the hypothesis that ``asking a model for answer A to question Q (generative)'' and ``asking a model if A is the answer to Q (discriminative)'' lack consistency~\cite{liang2023uhgeval}. They prompt the generator to produce both correct and incorrect answers, then use the discriminator to evaluate its own responses, aiming for the generator and discriminator to reach a Nash equilibrium. They select the best response based on the degree of consistency.

The significant drawback of this line of work is the high inference cost, as it often requires different LLM instances, potentially consuming multiple times the GPU memory and increasing the inference burden due to the extensive context generated by agents. Additionally, most models need a stopping condition to end the dialogue, and fixed round stopping is inflexible and can reduce performance. There is no current flexible and efficient stopping criterion. However, Multi-Agent systems remain a promising AI direction, and cost issues shouldn't deter exploration.

\section{Task: Hallucination Alleviation} \label{sec:hallucination_alleviation}

\noindent Hallucination alleviation is aimed at open-ended generation tasks such as story writing and code generation, emphasizing goals like fact enhancement, error reduction, and faithfulness enhancement. We have categorized four significant lines of work, as shown in the lower half of Table~\ref{tab:reason_hall_paradigms}.

\subsection{Refining the Response Iteratively} \label{sec:refining_the_response_iteratively}

\begin{figure}[t!]
    \centering
    \includegraphics[width=\linewidth]{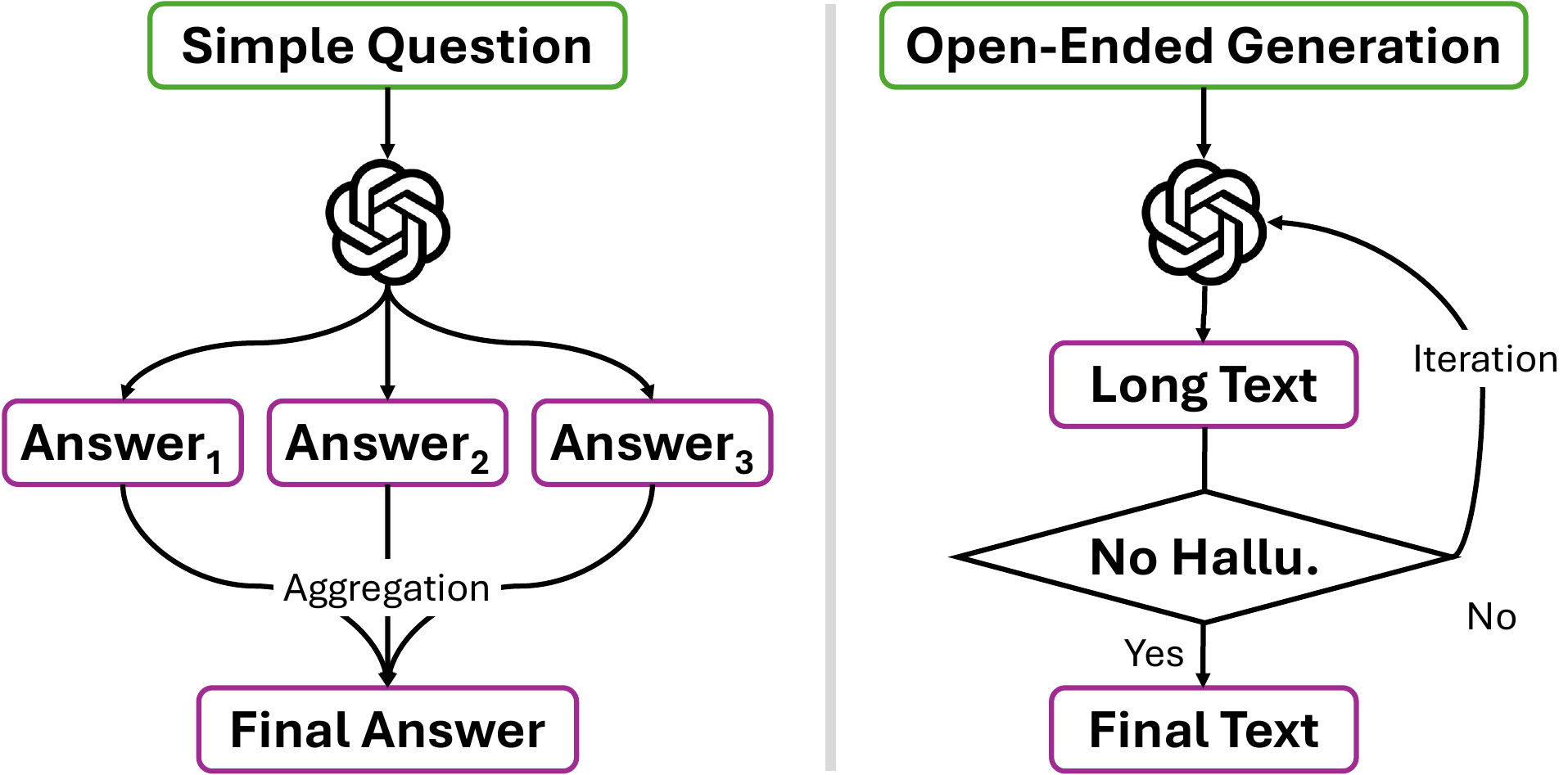}
    \caption{Refining with Responses (Left) V.S. Refining the Response Iter. (Right)}
    \label{fig:two_refine}
\end{figure}

\noindent This line of work is similar to Refining with Responses (Section~\ref{sec:refining_with_responses}) which primarily targets simple QA tasks. While Refining the Response Iteratively (Section~\ref{sec:refining_the_response_iteratively}) primarily deals with open-ended tasks such as story generation and code generation. Their comparison is shown in Fig.~\ref{fig:two_refine}.

The most famous works include Self-Refine~\cite{SelfRefine_23_NeuIPS_CMU}, Reflexion~\cite{Reflexion_23_NeuIPS_Northeastern}, and Self-Correct~\cite{SelfCorrect_23_ICLR_AI2}. These three frameworks share the basic structure of having the LLM provide textual feedback, which is then used to update the response iteratively until a stopping criterion is met or the maximum iterations is reached, as shown in Algorithm~\ref{alg:refining_the_response_iteratively}.

\begin{algorithm}[h]
\small
\caption{\textsc{Refining the Response Iteratively}}
\label{alg:refining_the_response_iteratively}
\begin{algorithmic}[1]
\REQUIRE Input query $\boldsymbol{x}$, model $\mathcal{M}$, consistency signal generator $\text{SelfEvaluate}(\cdot)$, Self-Update strategy $\text{SelfUpdate}(\cdot)$, stopping criterion $\text{stop}(\cdot)$, max iteration $T$
\STATE $\boldsymbol{y}_0 = \mathcal{M}(\boldsymbol{x})$
\STATE $i \gets 0$
\WHILE {i \textless T \AND \NOT $\text{stop}(\boldsymbol{y}_i)$}
    \STATE $f_i = \text{SelfEvaluate}(\boldsymbol{x}, \boldsymbol{y}_i)$
    \STATE $\boldsymbol{y}_{i+1} = \text{SelfUpdate}(\boldsymbol{x}, \boldsymbol{y}_{0:i}, f_{0:i})$
    \STATE $i \gets i + 1$
\ENDWHILE
\RETURN $\boldsymbol{y}_i$
\end{algorithmic}
\end{algorithm}

% We can also formalize the process in Algorithm~\ref{alg:refining_the_response_iteratively} as a conditional language model, as shown in Eq.~\ref{eq:refining_the_response_iteratively}.

% \begin{equation}
%     P(\boldsymbol{y}|\boldsymbol{x}) = P(\boldsymbol{y}_0|\boldsymbol{x}) \prod_{i=1}^{n} P(\boldsymbol{y}_i|\boldsymbol{y}_{i-1}, \boldsymbol{x}, f_{i-1})
%     \label{eq:refining_the_response_iteratively}
% \end{equation}

% Here, $P(\boldsymbol{y}_0|\boldsymbol{x})$ is a common conditional language model, $n$ denotes the number of iterations, $f_i$ is the Feedback Signal, and $P(\boldsymbol{y}_i|\boldsymbol{y}_{i-1}, \boldsymbol{x}, f_{i-1})$ represents the refinement based on the previous output $\boldsymbol{y}_{i-1}$ and the generated feedback $f_{i-1}$.

Despite following a similar framework, there are differences in specific implementations. Self-Refine~\cite{SelfRefine_23_NeuIPS_CMU} is the most naive implementation, where $\text{SelfEvaluate}(\cdot)$ is entirely performed by the LLM to generate textual feedback. Reflexion~\cite{Reflexion_23_NeuIPS_Northeastern} takes a better approach by viewing the iterative refining process as Verbal Reinforcement Learning, which is reinforcement learning without weight updates. Additionally, they separate feedback into feedback signal generation (e.g., error messages generated after code compilation in code generation tasks) and textual feedback generation (reflecting on error messages), increasing the framework's completeness. However, this approach requires a specific feedback signal design for each task, reducing its generality. Self-Correct~\cite{SelfCorrect_23_ICLR_AI2} uses the same framework but trains a dedicated Corrector model to generate better feedback. This method, however, is still not task-agnostic and significantly reduces the framework's flexibility due to the introduction of training.

The works mentioned above mainly construct frameworks for general tasks, while some focus on specific tasks. For example, Re$^3$~\cite{Re3_23_EMNLP_Berkeley} draws inspiration from human actions in writing long stories and proposes a draft, rewrite, and edit cycle to optimize the LLM's ability to write long stories. PEER~\cite{PEER_23_ICLR_Meta} mimics human collaborative editing by having the LLM iteratively propose editing suggestions to complete Wikipedia text editing. Self-Debug~\cite{SelfDebug_24_ICLR_Google} allows the model to debug its code through execution results and self-written unit test results, gradually refining the code until it is perfected.

\subsection{Mitigating Hallucination while Generating} \label{sec:mitigating_hallucination_while}

\noindent As mentioned earlier, hallucinations often manifest in finer details, such as temporal inaccuracies, date errors, or misattributions of names~\cite{liang2023uhgeval}. Multi-round iterations may overlook these minor errors, prompting some works to propose methods for more granular error editing, mitigating hallucination while generating\footnote{This section incorporates ideas from RAG, yet given its relevance to Self-Feedback, it's delineated as a distinct line of work.}. Currently, this is not yet a relatively mature direction, and there is no unified solution emerging. The following outlines typical approaches in methodology.

M\"undle et al.~\cite{HalluSelfContradictory_24_ICLR_ETH} utilize the phenomenon of Self-Contradiction to eliminate hallucinations\footnote{Demo of Self-Contradiction: \url{https://chatprotect.ai/}}. Specifically, it induces prompts to generate two contradictory sentences and then directs the LLM to resolve the contradictions, retaining the consistent information to generate a coherent sentence. Subsequent sentences follow a similar approach to produce a complete reply. Clearly, contradictory information is highly likely to be hallucinatory, thus effectively mitigating hallucinations. This method essentially extends Self-Consistency~\cite{SelfConsistency_23_ICLR_Google} into the domain of hallucination.

EVER (REal-Time VErification and Rectification)~\cite{EVER_arXiv_23_UNC} employs a similarly intuitive approach. When generating a sentence, EVER verifies the accuracy of the generated sentence either by the LLM itself or retrieved external information, generating feedback to modify the sentence if there are issues. The modified sentence is then re-appended into the generated text iteratively. Similarly, PURR (Petite Unsupervised Research and Revision)~\cite{PURR_23_arXiv_UCI} and RARR (Retrofit Attribution using Research and Revision)~\cite{RARR_23_ACL_CMU} follow a similar approach as EVER, where the verification stage relies on retrieving external knowledge to provide modification feedback.

In contrast to EVER, FAVA (FAct Vericaton with Augmentation)~\cite{FAVA_24_arXiv_Washington} adopts a more sophisticated approach. It fine-tunes the model to generate special tokens that edit its own content, enhancing editing efficiency\footnote{Their fine-tuning dataset includes examples like: ``Messi is an \textless entity\textgreater \textless delete\textgreater Argentine \textless /delete\textgreater \textless mark\textgreater Brazilian \textless /mark\textgreater \textless /entity \textgreater soccer player.'' Special tokens enclosed in angle brackets are also trained to be generated, effectively eliminating hallucinations through rendering.}. The major advantage of this method lies in granting the LLM maximum autonomy to make mistakes and subsequently correct them freely. Moreover, this approach bears resemblance to Quiet-STaR~\cite{QuietSTaR_24_arXiv_Stanford} mentioned in Section~\ref{sec:reasoning_topologically}, where both utilize special tokens to represent essential cognitive processes.

\subsection{Decoding Truthfully} \label{sec:decoding_truthfully}

\noindent Decoding Truthfully focuses predominantly on decoding consistency. In recent years, several studies have discovered that methods such as greedy decoding and sampling decoding constrain LLMs from accurately expressing crucial information in natural language. Consequently, more complex and rational decoding strategies have been designed to elevate the reliability and accuracy of model's responses~\cite{liang2024controlled}.

Li et al.~\cite{CD_22_arXiv_Stanford} pioneered the Contrastive Decoding strategy, where during the next token prediction, the optimal token probability is selected by contrasting the token probability distributions derived from expert and amateur models, as shown in Eq.~\ref{equ:ContrastiveDecode}. This method excels in mitigating biases or preferences inherent in large-scale models, favoring tokens with higher probabilities in expert models and lower probabilities in amateur models.

\begin{equation}
\label{equ:ContrastiveDecode}
\boldsymbol{y}^t \sim \operatorname{softmax}\left(\log \frac{P_{\mathrm{EXP}}\left(\boldsymbol{y}^t \mid \boldsymbol{y}^{0:t-1}\right)}{P_{\mathrm{AMA}}\left(\boldsymbol{y}^t \mid \boldsymbol{y}^{0:t-1}\right)}\right)
\end{equation}

Following this pioneering work, researchers have explored various approaches for logit adjustment and contrastive decoding. Chuang et al.~\cite{DoLa_24_ICLR_MIT} observed significant differences in token probability distributions across different layers of the model and introduced DoLa to incorporate information from previous layers, enhancing early-stage cognitive reasoning and pre-answer consistency, termed Decoding Consistency.

Unlike DoLa, SED~\cite{SED_24_arXiv_FDU} and DIVER ~\cite{DIVER_24_arXiv_IA} focus on detecting and addressing discrepancies caused by differences in tokens at certain positions, termed Chaotic Points. Methods for detecting chaotic points include comparing the ratio of maximum to second-maximum token probabilities or the number of candidate tokens exceeds one. Their indicator functions are shown in Eqs.~\ref{equ:chaotic_ratio} and ~\ref{equ:chaotic_base}, where $\delta_r$ is a probability threshold, $\gamma$ is a predefined coefficient, and $\mathcal{V}$ denotes the vocabulary. By assessing previously generated contents against potential tokens from chaotic points, scores such as information gain, weighted uncertainty, and weighted confidence help identify the most suitable token.

{ \footnotesize
\begin{equation}
\label{equ:chaotic_ratio}
\mathbb{I}_1 \left(\frac{P_{\text {second }}}{\boldsymbol{p}_{\text {max }}} \geq \delta_r\right)
\end{equation}

\begin{equation}\label{equ:chaotic_base}
\mathbb{I}_2 \left( \left| \left\{ \boldsymbol{y}^t \mid P\left( \boldsymbol{y}^t \mid \boldsymbol{y}^{0:t-1}\right) \geq \gamma \max_{w \in \mathcal{V}} P\left( w \mid \boldsymbol{y}^{0:t-1}\right) \right\} \right| > 1 \right)
\end{equation}
}

Those methodologies primarily apply to closed-book generation tasks. For open-book generation tasks, current research focuses on leveraging external references to guide decoding. CAD~\cite{ContextAwareD_23_arXiv_Washington} and ECAD~\cite{ContextDecode_24_arXiv_Edin} (named ECAD in this survey) incorporate contextually relevant or irrelevant knowledge snippets into model inputs, intervening in the decoding process through contrastive decoding strategies to bridge the information gap between useful and non-useful information.

\subsection{Activating Truthfulness} \label{sec:activate_truth}

\noindent  Activating Truthfulness focuses on enhancing consistency in latent layers. Its core methods involve boosting attention heads and states that represent ``truthfulness'' within latent layers, aiming to improve the model's internal consistency.

The exploration of latent truthfulness began with CCS (Contrast-Consistent Search)~\cite{CCS_23_ICLR_UCB}. CCS investigates methods for mining knowledge embedded in latent layers by training a small classification head on Transformer latent layers. This method effectively activates model truthfulness, surpassing conventional inference methods.

Inspired by CCS, Harvard scholars introduced the Inference-Time Intervention (ITI) technique~\cite{ITI_23_NeuIPS_Harvard}. ITI consists of two steps: 1) Probe analysis: Using probe technology\footnote{A probe is a small classifier whose input is latent states and whose output is labels corresponding to a test task.} to identify attention heads in the model related to truthfulness. 2) Inference-time intervention: The model's answer generation process is adjusted by increasing the weights of selected attention heads, guiding the model toward more truthful reasoning. However, ITI has limitations in training probes using only the last token's latent layer state at the end of a QA pair. TrFr~\cite{TrFr_24_AAAI_BUAA} addressed this by using multi-dimensional orthogonal probes to extract features from both truthful and non-truthful texts, improving attention head identification. TruthX~\cite{TruthX_24_ACL_ICT} explored a more efficient intervention strategy. It targets not only attention heads but also the feed-forward network layers. Mapping these states separately using truthful and semantic encoders significantly reduces the impact on the language model's overall performance while enhancing representations of truthfulness.

\textbf{White-Box Hallucination Alleviation.} Mitigating hallucinations from a white-box perspective involves activating the internal authenticity of the model, which necessitates interpretability studies. For instance, a recent survey~\cite{zheng2024attention} reveals that attention heads in models can serve various functions. Building on these functional distinctions, we may discover better approaches to mitigate hallucinations. For example, Wu et al.~\cite{RetrievalHead_24_arXiv_PKU} found that certain attention heads are more adept at long-context retrieval (strong ``copy-paste'' abilities). In tests such as Needle-in-a-Haystack, blocking these attention heads caused performance to drop from 94.7\% to 63.6\%. Can enhancing retrieval heads reduce hallucinations in long contexts? This is a question worth investigating.

\section{Task: Others} \label{sec:other_tasks}

\noindent Several works follow the Self-Feedback framework, though not always targeting internal consistency. For completeness, we summarize these efforts below.

\subsection{Preference Learning} \label{subsec:PL}

\noindent Preference Learning (PL) aims to align LLM outputs with human intent~\cite{wu2024self, pmlr-v235-chen24j, pmlr-v235-pang24a}. Most of the work around this task can be broadly covered by the Self-Feedback framework. For PL, the Feedback Signal mainly refers to the reward information given by a reward model $\mathcal{R}$, which is trained through preference feedback. Preference feedback involves comparing and ranking different responses to the same question in terms of helpfulness, harmlessness, and honesty. The Self-Update here primarily refers to broadly updating the model $\mathcal{M}$, including methods like supervised fine-tuning and reinforcement learning (such as PPO~\cite{PPO_17_arXiv_OpenAI}, DPO~\cite{DPO_23_NIPS_Stanford}).

There are three main ways to obtain preference feedback. 1) Through human feedback, as seen in works like OASST~\cite{OASST_23_NIPS_XX} and BeaverTails~\cite{BeaverTails_23_NIPS_PKU}, which include human-annotated data. 2) Feedback generated by models~\cite{ConstitutionalAI_22_arXiv_anthropic,SALMON_24_ICLR_IBM}, offering lower annotation costs and faster iterative feedback efficiency compared to human feedback. 3) Feedback derived from inductive bias, such as upvotes/downvotes in the SHP dataset~\cite{DataDifficult_22_PMLR_UoW}, or prior rules in ALMoST~\cite{ALMoST_23_arXiv_NAVER}, which rank response quality based on model size or prompt context.

Based on preference feedback, we can train a reward model to output Feedback Signals. There are two common types of reward models. One is the Reward Model proposed in InstructGPT~\cite{InstructGPT}, with the loss function as shown in Eq.~\ref{equ:reward_model}. Here, $r_\theta(\boldsymbol{x}, \boldsymbol{y})$ represents the output of the Reward Model, and response $\boldsymbol{y}_w$ is ranked higher than $\boldsymbol{y}_l$. However, this method's downside is that the overall score distribution for high-quality and low-quality responses is similar, making it difficult to effectively distinguish between different responses to different questions. To address this, Xu et al.~\cite{SelfCritique_24_arXiv_Zhipu} proposed an evaluation model that directly scores QA pairs.

\begin{equation}\label{equ:reward_model}
\begin{aligned}
z &= \sigma\left(r_\theta\left(\boldsymbol{x}, \boldsymbol{y}_w\right) - r_\theta\left(\boldsymbol{x}, \boldsymbol{y}_l\right)\right) \\
\operatorname{loss}(\theta) &= -\frac{1}{\binom{k}{2}} E_{\left(\boldsymbol{x}, \boldsymbol{y}_w, \boldsymbol{y}_l\right) \sim D}\left[\log \left(z\right)\right]
\end{aligned}
\end{equation}

% 上面提到的工作在Self-Update阶段都是重新推理，与其不同，RLAIF~\cite{ConstitutionalAI_22_arXiv_anthropic}使用AI反馈进行强化学习，其中教师模型对学生模型的输出进行排名，得到一个数据集，以此为Feedback Signal，并在Self-Update阶段使用得到的数据集来训练学生模型。
% 类似地，ReST算法~\cite{ReST_23_arXiv_Google}利用学习得到的奖励模型对学生产生的多个输出进行排名，构建数据集。

\subsection{LLM-Based Knowledge Distillation} \label{subsec:KD}

\noindent LLM-based knowledge distillation methods aim to transfer advanced capabilities from proprietary LLMs (such as GPT-4) to small-parameter open-source models~\cite{SurveyKD_24_arXiv_HKU}. These two models can be referred to as the ``teacher model'' and the ``student model'' respectively, with the teacher model guiding the student model to enhance its capabilities, fitting the generalized Self-Feedback framework proposed in this paper. During the Self-Evaluation, the student model generates answers, which are then assessed by the teacher model. In the Self-Update, the student model uses the evaluation signal to update itself or its answers.

This signal can be in the form of statistical metrics, such as MiniLLM~\cite{MiniLLM_24_ICLR_THU} calculating the reverse Kullback-Leibler (KL) divergence of the probability distributions output by the student and teacher models; or GKD~\cite{GKD_24_ICLR_Google} computing metrics like forward KL divergence, reverse KL divergence, and generalized JSD. The signal can also be textual feedback, such as Selfee~\cite{Selfee_23_blog} utilizing ChatGPT as the teacher to provide textual feedback on the outputs of the student model; or in PERsD~\cite{PERsD_23_EMNLP_NTU}, where the teacher executes the code generated by the student model and provides specific suggestions based on errors.

% FIGA~\cite{FIGA_24_ICLR_RUC}中教师根据学生的响应和golden response进行对比并据此反馈修改意见。
% 另外，还有研究者使用教师模型对学生模型生成的token概率分布进行评价反馈，然后学生模型结合反馈信息进行梯度更新。如

When the teacher and student models are the same LLM, this leads to Self-Knowledge Distillation (Self-KD). In Self-KD, the model iteratively updates its capabilities using the knowledge it gradually accumulates during training, falling under the narrow Self-Feedback paradigm. For example, the goal of Impossible distillation~\cite{ImposDistill_23_arXiv_UoW} is to obtain a Stronger Paraphraser. In the Self-knowledge distillation process, it evaluates its paraphrase results from perspectives such as semantics, format, and diversity, and further refines high-quality data to fine-tune itself accordingly.

% Hahn et al.~\cite{SelfKnowledge_19_RANLP_Handong}观察到在语义空间中相近的token应该有相似的logit值，于是设计在训练过程中获取并利用token的嵌入向量修正对应token的目标概率，然后利用这个修正后的概率计算损失函数并进行梯度更新。
% 类似地，~\cite{TFKD_20_CVPR_NUS}采用了Label Smoothing Regularization的方法将目标标签转化为软概率标签。~\cite{PSKD_21_ICCV_LG}用上一轮训练得到的checkpoint推理得到一个概率分布，将这个分布和目标标签加权得到软概率标签。

\subsection{Data Augmentation}

\noindent Data Augmentation aims to construct and filter high-quality datasets using LLMs. It is somewhat similar to the methods in Sections \ref{subsec:PL} and \ref{subsec:KD} that combine Feedback information to create datasets, but there are slight differences in focus and specific forms. The latter focuses on the model's capabilities, using datasets during the Self-Update stage for model fine-tuning, with most methods falling under narrow Self-Feedback. In contrast, Data Augmentation focuses on the dataset itself, updating the model's responses during the Self-Update stage to further refine the dataset, with most methods falling under generalized Self-Feedback.

Self-instruct~\cite{SelfInstruct_23_ACL_Washington} is a typical example, where the LLM generates new task instructions during the Self-Evaluation stage and generates input-output instances based on the new instructions. It calculates the ROUGE-L metric between the new instructions and existing instructions as the Feedback signal. Finally, during the Self-Update stage, it filters and screens the newly generated set of instructions.

Currently, methods applying LLMs to Data Augmentation and Synthetic Data Generation mainly focus on the prompt engineering layer. In other words, Self-Evaluation only involves responses. Many studies have shown that LLM responses are highly sensitive to prompt variations~\cite{PromptSensitive_23_arXiv_UoW,yu2024xfinder}. Therefore, the main bottleneck in this task is: how to design better prompts and how to deeply explore the relationship between decoding, latent states, and data quality.

% \subsection{}
% 持续学习
% TODO：再列举新方向有必要，但优先级低

\section{Evaluation} \label{sec:evaluation}

\noindent This section covers evaluation methods and benchmarks for internal consistency and Self-Feedback, focusing on two abilities: meta (e.g., uncertainty, consistency, feedback) and common (e.g., reasoning QA, code generation) abilities. Meta evaluation identifies which LLMs are the best, while common evaluation reveals which Self-Feedback methods are the best.

\subsection{Meta Evaluation}

\noindent We summarize five meta evaluation methods, categorized into \iconruler{}metric-based and \icondata{}benchmark-based approaches. Metric-based methods calculate performance mainly via formulas, while benchmark-based methods empirically measure it using QA datasets (see Table~\ref{tab:meta_eval}).

\begin{table}[h!]
\centering
\caption{Meta Evaluation Benchmarks} \label{tab:meta_eval}
\begin{tabular}{lll}
\toprule
\textbf{Type} & \textbf{Benchmark} & \textbf{Organization} \\
\midrule
Uncertainty & LLM-Uncertainty-Bench~\cite{UncertaintyBench_24_arXiv_Tencent} & Tencent \\
Uncertainty & UBench~\cite{UBench_24_arXiv_Nankai} & Nankai \\
Consistency & ConsisEval~\cite{ConsisEval_24_arXiv_PKU} & PKU \\
Consistency & PopQA-TP~\cite{PopQA_23_GEM_IBM} & IBM \\
Consistency & ParaRel~\cite{ParaRel_21_TACL_BarIlan} & BIU \\
Consistency & BMLAMA~\cite{CrossLingualConsistency_23_EMNLP_UoGroningen} & RUG \\
Consistency & BECEL~\cite{BECEL_22_Coling_Oxford} & Oxford \\
Critique Ability & CriticBench~\cite{CriticBench_24_arXiv_THU} & THU \\
Self-Knowledge & SelfAware~\cite{TheoryKnowUnknown_23_ACL_Fudan} & Fudan \\
Self-Knowledge & Idk(I don't know)~\cite{TheoryKnowUnknown_24_arxiv_Fudan} & Fudan \\
Self-Knowledge &  Self-Knowledge Evaluation~\cite{EvalSelf_24_arXiv_THU} & THU \\
\bottomrule
\end{tabular}
\end{table}

\textbf{\iconruler{}Uncertainty Evaluation\footnote{As mentioned in Section~\ref{sec:uncertainty}, uncertainty estimation involves assessing the uncertainty of a model's specific response. Uncertainty evaluation, on the other hand, measures the overall uncertainty of a model.}.} Key metrics for evaluating model uncertainty include: Expected Calibration Error (ECE), which assesses the expected difference between model confidence and accuracy; Maximal Calibration Error (MCE), which indicates the maximum deviation between model accuracy and confidence; and Brier Score (BS), which is used to assess how closely the model’s predicted probabilities align with the true class probabilities~\cite{SurveyUncertainty_23_arXiv_Nankai}.

\textbf{\icondata{}Uncertainty Evaluation.} LLM-Uncertainty-Bench~\cite{UncertaintyBench_24_arXiv_Tencent} extracts five test tasks (including question answering, reading comprehension, commonsense inference, dialogue response selection, and document summarization) from common benchmark datasets and uses conformal prediction techniques to construct benchmarks. UBench~\cite{UBench_24_arXiv_Nankai} also extracts data from other datasets, totaling 3978 multiple-choice questions covering knowledge, language, understanding, and reasoning abilities. UBench evaluates individual data items by having models textually express uncertainty scores.

\textbf{\icondata{}Consistency Evaluation.} This line of work centers on assessing whether a model delivers consistent responses to queries that are semantically equivalent but phrased differently. The key focus is on developing a variety of synonymous queries to test the model's reliability. For instance, the ConsisEval Benchmark~\cite{ConsisEval_24_arXiv_PKU} creates simpler synonymous queries for each question. PopQA-TP~\cite{PopQA_23_GEM_IBM} and ParaRel~\cite{ParaRel_21_TACL_BarIlan} construct synonymous queries through rephrasing. BMLAMA~\cite{CrossLingualConsistency_23_EMNLP_UoGroningen} focuses on multilingual consistency, constructing a parallel corpus of queries. BECEL~\cite{BECEL_22_Coling_Oxford} draws inspiration from behavioral consistency, considering higher-order consistency in model responses by creating semantic consistency data, negational consistency data, symmetric consistency data, etc. Notably, most studies have found that models generally exhibit low consistency.

\textbf{\icondata{}Critique Abilitiy Evaluation.} Lin et al.~\cite{CriticBench_24_arXiv_THU} collect a large number of QA pairs from 15 datasets across mathematical, commonsense, symbolic, coding, and algorithmic fields, creating CriticBench through model generation and human annotation. It can be used to evaluate the ability of LLMs to generate critiques, an important aspect of the Self-Feedback framework.

\textbf{\icondata{}Self-Knowledge Evaluation.} Self-Knowledge refers to the LLM's understanding and recognition of its own abilities, limitations, and the content it creates. Yin et al.~\cite{TheoryKnowUnknown_23_ACL_Fudan} and Cheng et al.~\cite{TheoryKnowUnknown_24_arxiv_Fudan} construct sets of unanswerable questions to explore the question ``Do large language models know what they do not know?'' Tan et al.~\cite{EvalSelf_24_arXiv_THU} investigate ``Does the model truly understand the questions and solutions it creates?'' These studies generally yield negative empirical results, indicating that models have weak Self-Knowledge.

\subsection{Common Evaluation}

\noindent Self-Feedback methods are often evaluated using benchmarks that focus on real-world tasks like reasoning, code generation, and math problem solving (see Table~\ref{tab:benchmarks}). For more information on LLM evaluation, you can refer to this survey~\cite{surveyllmeval}.

\begin{table}[h!]
\centering
\caption{Common Evaluation Benchmarks}
\label{tab:benchmarks}

\begin{tabular}{lll}
\toprule
\textbf{Type} & \textbf{Benchmark} & \textbf{Organization} \\
\midrule
Knowledge reasoning                          & C-Eval~\cite{CEval}                          & SJTU                                    \\
Knowledge reasoning                          & MMLU~\cite{hendrycks2021measuring}     & UCB                                  \\
Logic reasoning                              & BBH~\cite{Bench_BBH}                         & Google                                   \\
Logic reasoning                              & ARC~\cite{ARC_c}                             & AI2                                  \\
Linguistic understanding                     & WiC~\cite{bench_WiC}                         & Cambridge                                   \\
Code generating                              & HumanEval~\cite{HumanEval}                   & N/A                                  \\
Math Solving                                        & MATH~\cite{bench_math}                       & UCB                                   \\
Math Solving    & GSM8K~\cite{cobbe2021training}               & OpenAI                                  \\ 
\bottomrule
\end{tabular}

\end{table}

% Currently, these Benchmark evaluation formats mainly include multiple-choice question answering (MCQA) and text generation, each with its own advantages and disadvantages. For MCQA, evaluators can accurately calculate metrics such as accuracy by extracting answer text from the response or comparing the probabilities of different option tokens during LLM decoding. However, many researchers pointed out the drawbacks of this method, namely that LLMs tend to choose the first option, leading to poor self-consistency and not reflecting the LLM's true ability~\cite{FTEvsTOE_24_arxiv_LMU,wang2024answersreviewingrationalitymultiple}. Text generation tasks, such as math problems and code generation, mitigate the shortcomings of MCQA by allowing the LLM to perform freely, better reflecting the model's true ability. However, this evaluation method struggles to precisely quantify the gap between generated text and reference answers in dimensions such as syntax and semantics. It is evident that current Benchmarks for common evaluation are still far from accurately measuring the true capabilities of LLMs and have many areas for improvement.

\section{Does Self-Feedback Really Work?}  \label{sec:does_it_work}

\subsection{Conflicting Viewpoints} \label{sec:conflict}

\noindent With the rise of works prefixed by ``Self-'', questions of feasibility arise: Can a model truly optimize itself? Many studies have attempted to answer this question, with most focusing on Refining the Response Iteratively and Multi-Agent Collaboration.

\begin{itemize}
    \item Jiang et al.~\cite{SelfIncorrect_24_arXiv_JHU} propose the SELF-[IN]CORRECT hypothesis, showing that in QA tasks, models are better at generating answers than judging their own correctness, highlighting a self-assessment limitation.

    \item Stechly et al.~\cite{GPT4Doesnt_23_NeurIPS_ASU} and Valmeekam et al.~\cite{CanSelfCritique_23_NeurIPS_ASU} found GPT-4 fails to verify its solutions in the Graph Coloring and planning tasks, with verifiers generating many false positives, reducing reliability.

    \item Huang et al.~\cite{TheoryNoReason_24_ICLR_Google} refute the effectiveness of Reflexion~\cite{Reflexion_23_NeuIPS_Northeastern}, Multi-Agent Debate~\cite{Debate_23_arXiv_MIT}, and Self-Refine~\cite{SelfRefine_23_NeuIPS_CMU}. They argue Reflexion's reliance on external truth for refining is impractical, Multi-Agent Debate is inferior to Self-Consistency and resource-heavy, and Self-Refine's prompts were unfair, with better one-shot responses achievable through improved prompting.

    \item Kamoi et al.~\cite{SurveySelfCorrection_24_arXiv_PSU} provide a more comprehensive analysis by classifying various methods clearly and systematically comparing the strengths and weaknesses of each methods. They suggest that the ability to self-correct should be discussed according to the specific task. For example, for decomposable tasks\footnote{For example, ``Who are some politicians who were born in Boston?''} or verifiable tasks\footnote{For example, in the Game of 24 (Find arithmetic operations to obtain 24 using four given integers), generating a solution is harder than verification.}, it is feasible for the model to optimize itself.
\end{itemize}

While these criticisms reveal certain limitations in feedback signals, experimental tasks, and test models, they can be seen as limited perspectives~\cite{SelfIncorrect_24_arXiv_JHU, GPT4Doesnt_23_NeurIPS_ASU, CanSelfCritique_23_NeurIPS_ASU, TheoryNoReason_24_ICLR_Google}. Although the survey~\cite{SurveySelfCorrection_24_arXiv_PSU} provides more meaningful viewpoints through classified discussions, it complicates the field, making it difficult to form a systematic framework. Benefiting from the perspective of internal consistency and the clear boundary discussions in Section~\ref{sec:out_of_scope}, we conduct a more meaningful discussion on the proposed Self-Feedback framework: 

\begin{enumerate}
    \item \textbf{Does Self-Feedback improve internal consistency?} The answer is yes. As demonstrated in our survey, different lines of research offer affirmative evidence from various perspectives.
    \item \textbf{Does internal consistency mean correctness?} We cannot directly conclude this. We will delve deeper into this question in the following section.
\end{enumerate}

\subsection{Does Internal Consistency Mean Correctness?}

\begin{figure}[h!]
    \centering
    \includegraphics[width=\linewidth]{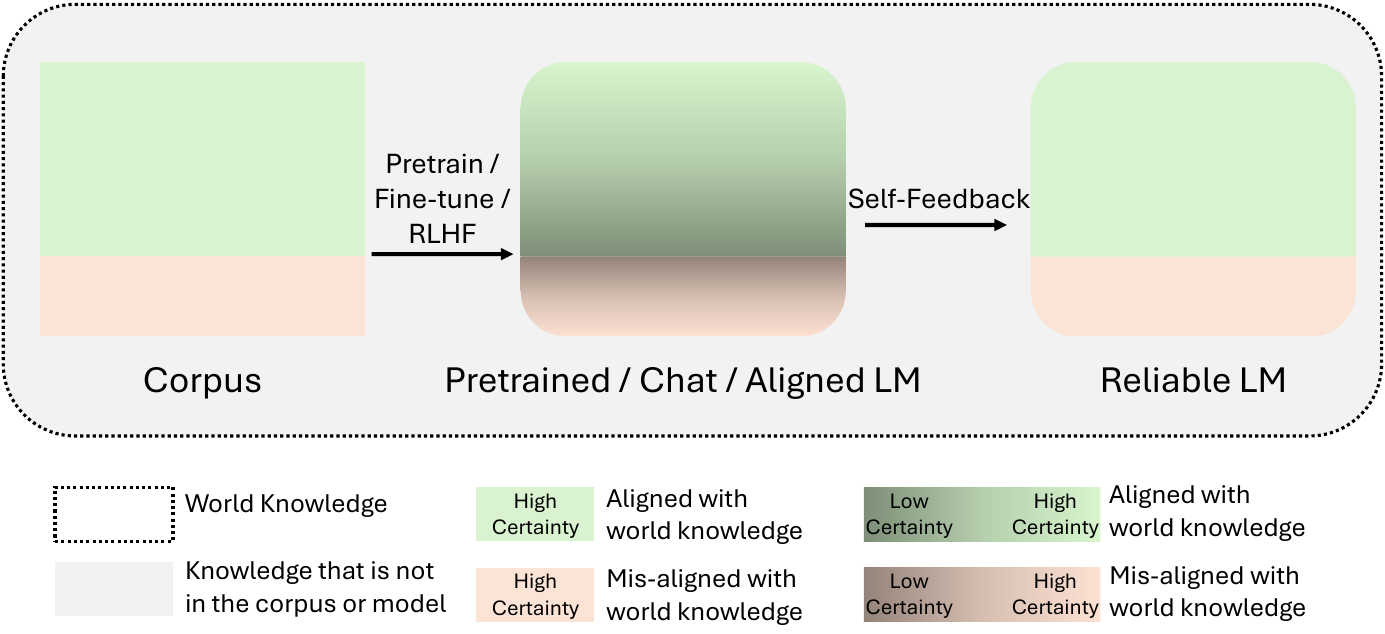}
    \caption{World Knowledge, Training Corpus and Language Model}
    \label{fig:knowledge}
\end{figure}

\noindent Let's revisit the relationship between world knowledge, training corpus, and language models (LMs), as shown in Fig.~\ref{fig:knowledge}. World knowledge is the consensual (correct) knowledge we humans possess. The training corpus used for models is a true subset of world knowledge, containing the vast majority of correct knowledge and a small portion of uncleanable erroneous knowledge. Additionally, the knowledge embedded in the corpus is deterministic, where each statement in the corpus has a probability of 100\%. Language models, by fitting the corpus, acquire higher-order probabilistic representations of this knowledge, but the probabilistic nature makes the learned knowledge vague and non-deterministic, as illustrated by the shaded areas in Fig.~\ref{fig:knowledge}. Vagueness (or hallucination) is an important characteristic of language models. It enables the generation of novel and creative expressions outside the training corpus distribution. However, from a reliability perspective, vagueness is a disaster. Vagueness means that answers to the same question are uncertain, making the model's expressions inconsistent.

Therefore, we need to improve internal consistency and eliminate vagueness within the model to enhance its confidence in correct knowledge. However, eliminating vagueness also means that the model will be equally confident in erroneous knowledge. This raises a question: does enhancing consistency yield overall benefits or drawbacks? The advantage is that when preprocessing and cleaning the pre-training corpus, the intention is to align it towards world knowledge (correct knowledge). Hence, we propose the ``Consistency Is (Almost) Correctness'' hypothesis.

\begin{tcolorbox}[colback=white!98!black,colframe=white!30!black,boxsep=1.1pt,top=6pt]
\textbf{Consistency Is (Almost) Correctness}\\[-0.575em]
\noindent\makebox[\textwidth]{\rule{\textwidth}{0.4pt}}
\\[0.25em]
Enhancing a language model's internal consistency activates its cognitive certainty, reinforcing both correct and erroneous knowledge. However, because the pre-training corpus is predominantly aligned with correct world knowledge, improving consistency tends to amplify correct content more than incorrect content. Consequently, increased internal consistency generally results in improved overall correctness.
\end{tcolorbox}

However, why do some opposing voices believe that improving consistency cannot enhance the model's correctness? We believe this is closely related to the testing tasks. Many works refuting Self-Feedback use testing tasks that lie in the shaded areas of Fig.~\ref{fig:knowledge} (e.g., unstated puzzles not in the training corpus or questions unsolvable without external knowledge). Models struggle to effectively Self-Evaluate and Self-Update for tasks beyond their generalization capability.

In summary, within-distribution capabilities, the Self-Feedback framework can enhance model consistency by reinforcing the model's fit to corpus priors, thereby eliminating uncertainty and improving consistency. According to the ``Consistency Is (Almost) Correctness'' hypothesis, this leads to an overall improvement in the model's performance.

% \subsection{Practical Suggestions}
% TODO 低优先级

% 既然有许多的争议，为了帮助大家更好地认识哪些方向对解决哪些任务更有帮助，我们提出一个技术路线图供大家参考，如图~\ref{fig:tech_map}所示。

% \begin{figure}[]
%     \centering
%     \includegraphics[width=\linewidth]{figures/placeholder.png}
%     \caption{技术路线图}
%     \label{fig:tech_map}
% \end{figure}

\subsection{Appeals}

\noindent The field faces significant criticism due to inconsistent naming, unrealistic tasks, varying benchmarks, and contradictory baselines. Thus, we propose the following appeals:

\begin{itemize}
    \item \textbf{Naming.} Ensure method names are distinct (e.g., Self-Improve~\cite{SelfImprove_23_EMNLP_Illinois} and Self-Improvement~\cite{SelfImprovement_24_arXiv_Tencent} are bad names) and accurate (e.g., uncertainty or confidence estimation).

    \item \textbf{Task Definition.} Standardize terms by adopting "Internal Consistency Mining" for reasoning elevation and hallucination alleviation tasks.

    \item \textbf{Reasoning and Hallucination.} Use ``lack of reasoning'' for QA tasks and ``exhibiting hallucination'' for open-ended generation tasks.
    
    \item \textbf{Selection of Baselines.} Select baselines from the same sub-direction (section) to ensure fair comparisons.

    \item \textbf{Experiment Settings.} Avoid unrealistic setups, such as requiring pre-given golden labels~\cite{TheoryNoReason_24_ICLR_Google}.

    \item \textbf{Prompt Engineering.} Disclose and test prompt templates for robustness and generality across different LLMs.
\end{itemize}

\section{Future Directions and Challenges} \label{sec:future}

\subsection{Textual Self-Awareness}

\noindent Human speech often lacks consistency and certainty in expressing viewpoints. However, we typically use phrases like ``I'm not sure, but I think'' or ``I believe there's an 80\% chance'' to hedge, demonstrating our good self-awareness. Yona et al.~\cite{ExpressUncertainty_24_arXiv_TAU} proved that current models still cannot verbally and faithfully express their uncertainty. Kapoor et al.~\cite{MustTaught_24_arXiv_NYU} found similar issues and showed through experiments that models can achieve good calibration only after fine-tuning. How to enable models to utilize the available internal consistency signal to help textually express their self-awareness is a promising direction~\cite{ExpressKnowledge_24_arXiv_FDU}.

\subsection{The Reasoning Paradox}

\noindent As mentioned in Section~\ref{sec:hourglass}, there is a paradox between the reasoning done during single token prediction (latent reasoning~\cite{TheoryLatentReason_24_arXiv_Google}) and the reasoning done using multiple tokens in language (explicit reasoning, e.g., CoT)~\cite{jin-etal-2024-impact}.

Therefore, we need to study the equilibrium point between latent and explicit reasoning, enabling efficient use of reasoning resources and improving the model's reasoning efficiency. Currently, there is little research on this issue.

\begin{tcolorbox}[colback=white!98!black,colframe=white!30!black,boxsep=1.1pt,top=6pt]
\textbf{The Paradox of Latent and Explicit Reasoning}\\[-0.575em]
\noindent\makebox[\textwidth]{\rule{\textwidth}{0.4pt}}
\\[0.25em]
Language models excel in latent reasoning when decoding a single token, effectively utilizing attention mechanisms and deep feature interactions to achieve accurate reasoning. However, single tokens can't answer complex questions. Explicit reasoning, which involves generating a sequence of tokens (e.g. CoT), enhances the model's problem-solving capabilities. Yet, lengthy reasoning chains and inherent noise in text disrupt the model's latent reasoning. Thus, there is a paradox between latent reasoning and explicit reasoning.
\end{tcolorbox}
% TODO：可以考虑增加示意图

\subsection{Dive Deeper}

\noindent From the seven lines of work we summarized, many works optimize only at the response layer. However, this approach relies on experience and is highly sensitive to prompt templates. Moreover, the low entry barrier and extensive participation in such work have led to an influx of low-quality papers. Therefore, we encourage researchers to delve into the decoding layer and latent layer, exploring more universal discoveries from an interpretability perspective.

\subsection{The Unified Perspective}

\noindent At present, the focus of work in this field is relatively narrow, lacking a comprehensive understanding of the entire field, and consequently, there are no more general framework works. We believe that using the perspective proposed in this paper, considering problems from the response, decoding, and latent layers in a unified manner, can better facilitate Internal Consistency Mining. There are emerging efforts that begin to integrate multiple layers. For example, Xie et al.~\cite{CalibIC_24_arXiv_SJTU} start from the response layer and reflect on how different CoT paths guide the consistency of the latent layer; Xie et al.~\cite{SelfEvaluation_23_NeuIPS_NUS} use Self-Evaluation strategies at the response layer to guide better decoding strategies.

\subsection{The Comprehensive Evaluation}

\noindent Different LLMs, combined with various Self-Feedback strategies, can produce vastly different combinations. However, as explained in Section~\ref{sec:evaluation}, current evaluation methods generally have a singular focus, making it difficult to comprehensively and conveniently understand the model's capabilities. Therefore, building a complete evaluation system from meta evaluation to common evaluation, from latent states to response, from benchmark to metric, and from uncertainty to feedback is a worthy consideration.

% TODO 暂不重要
% \subsection{Internal Consistency 建模}

\section{Conclusion}

\noindent This paper proposes using an internal consistency perspective to observe the most prominent phenomena in the field of LLMs: lack of reasoning and presence of hallucinations. The article explains the modeling of internal consistency, the hourglass evolution pattern, the current status, sources, and significance from multiple aspects, and proposes the Self-Feedback framework for Internal Consistency Mining. We summarize the various tasks and distinctive lines of work involved in the Self-Feedback framework. These lines of work can help researchers locate their work's position within a vast system and facilitate reasonable experimental comparisons. Finally, we include three critical topics: relevant evaluation methods and benchmarks, exploring whether Self-Feedback truly works, and future research directions. In summary, this paper attempts to use a deeper research perspective (Internal Consistency) and a more general framework (Self-Feedback) to summarize a series of important works on reasoning elevation and hallucination alleviation.

\section*{Acknowledgments}

% 短版本
% \noindent This work was supported by the National Natural Science Foundation of China (No. 62072463, 71531012) and the National Social Science Foundation of China (No. 18ZDA309).

% 长版本
\noindent This work was supported by the National Natural Science Foundation of China (Grants No. 62072463, 71531012), the National Social Science Foundation of China (Grants No. 18ZDA309), and the Research Seed Funds of the School of Interdisciplinary Studies at Renmin University of China.

% \balance
\bibliographystyle{IEEEtran}
\bibliography{IEEEabrv,references}

\begin{IEEEbiography}[{\includegraphics[width=1in,height=1.25in,clip,keepaspectratio]{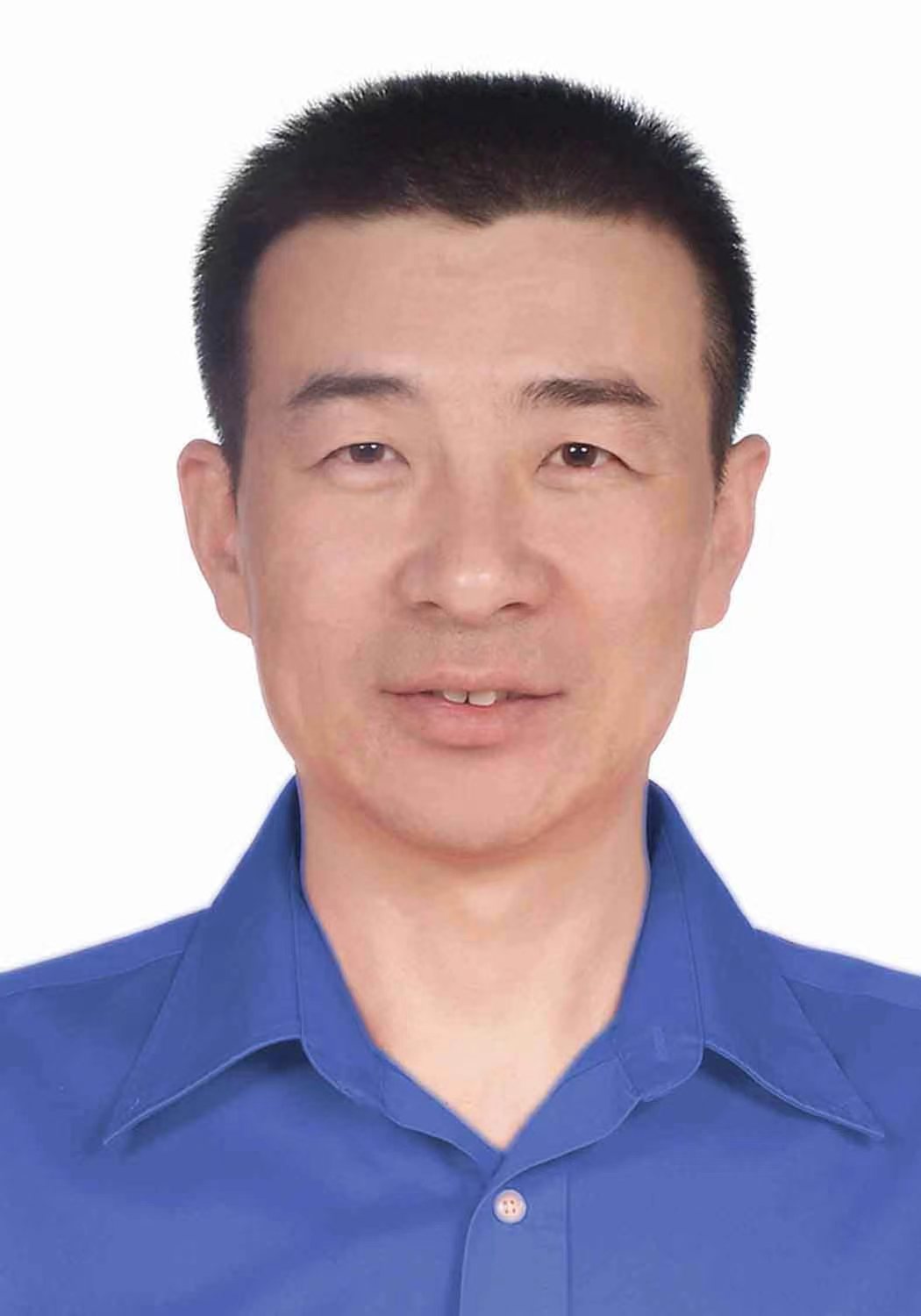}}]{Xun Liang}
(Senior Member, IEEE) received the B.Sc. and Ph.D. degrees in computer engineering from Tsinghua University, Beijing, China, in 1989 and 1993, respectively, and the M.Sc. degree in operations research from Stanford University, Palo Alto, CA, USA, in 1999. He worked as a Post-Doctoral Fellow with the Institute of Computer Science and Technology, Peking University, Beijing, from 1993 to 1995, and with the Department of Computer Engineering, University of New Brunswick, Fredericton, NB, Canada, from 1995 to 1997. He worked as a CTO, leading over ten intelligent information products in RixoInfo Ltd., CA, USA, from 2000 to 2007, and was the Director of the Data Mining Lab, Institute of Computer Science and Technology, Peking University, from 2005 to 2009. He is currently a professor with the School of Information, Renmin University of China. His research interests include support vector machines, social computing and large language models.
\end{IEEEbiography}

\vskip -2\baselineskip plus -1fil

\begin{IEEEbiography}[{\includegraphics[width=1in,height=1.25in,clip,keepaspectratio]{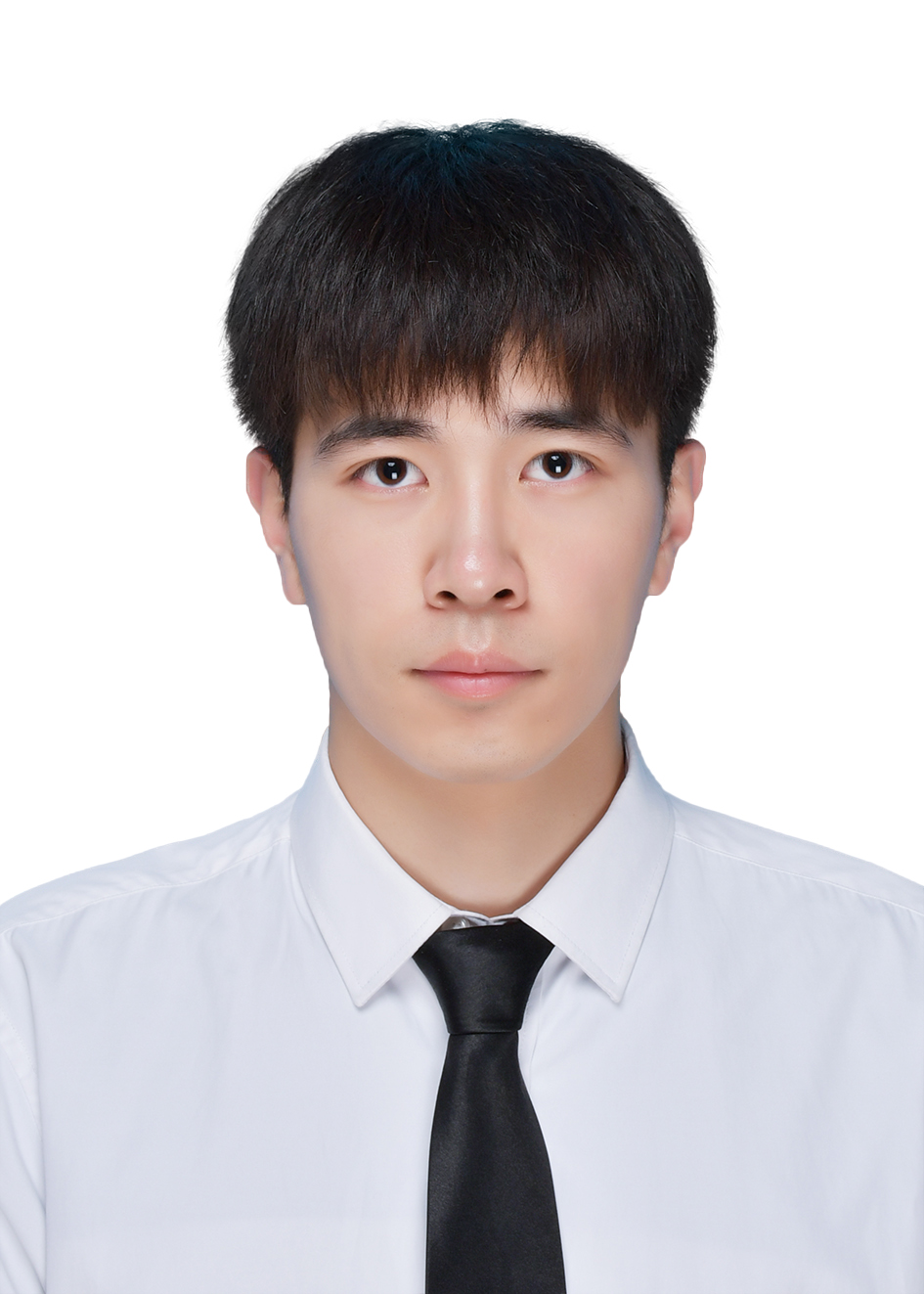}}]{Shichao Song}
is currently a PhD student at the School of Information, Renmin University of China, under the supervision of Prof. Xun Liang. His research interests span a wide range of topics, including internal consistency mining of LLMs, LLM interpretability, and reliable evaluation methods for LLMs. For more information, visit his website at \url{https://ki-seki.github.io/}.
\end{IEEEbiography}

\vskip -2\baselineskip plus -1fil

\begin{IEEEbiography}[{\includegraphics[width=1in,height=1.25in,clip,keepaspectratio]{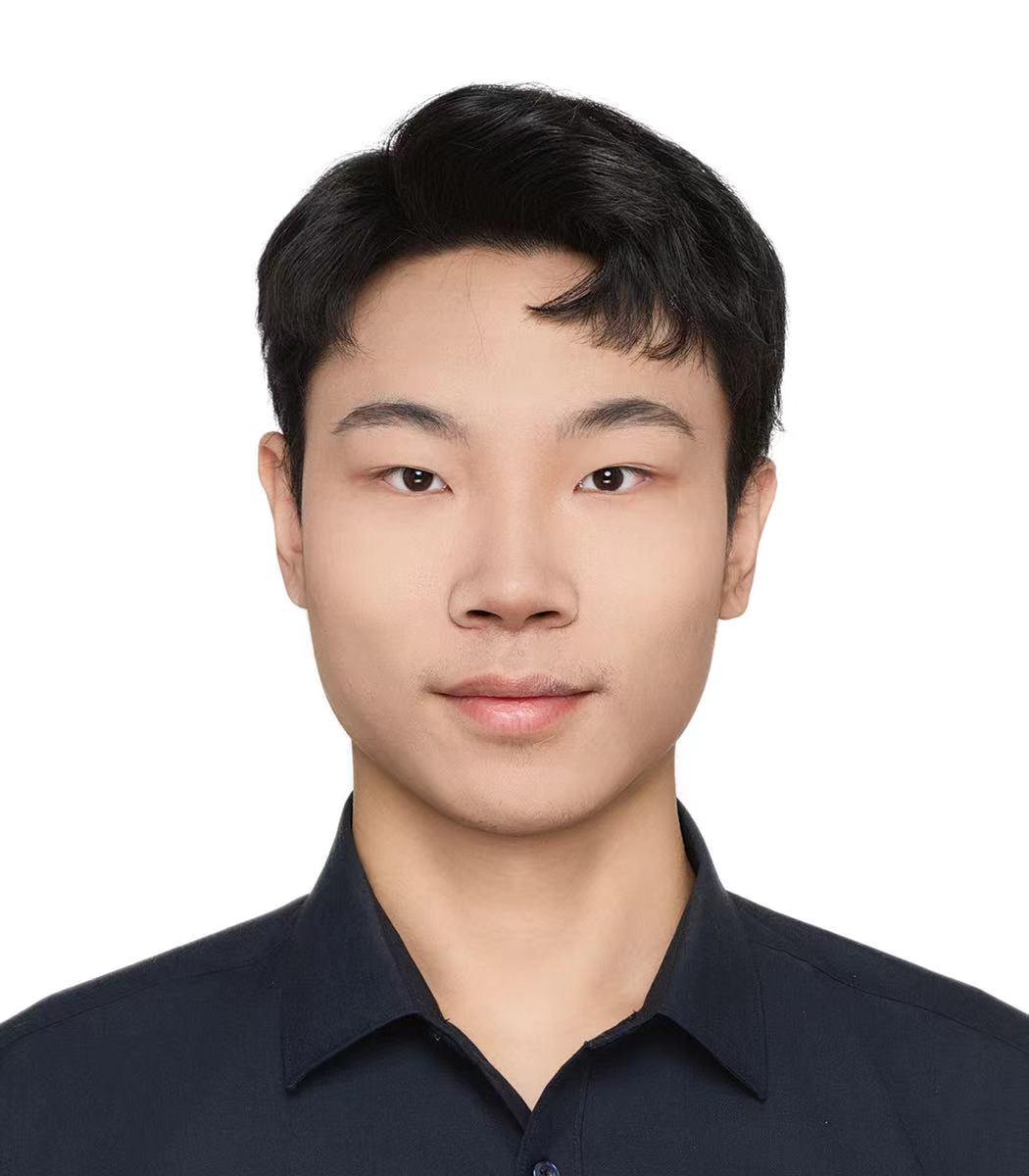}}]{Zifan Zheng}
is currently a research intern at the Large Language Model Center of the Institute for Advanced Algorithms Research, Shanghai. He received the B.S. degree in Computer Science and Technology from Beijing Institute of Technology, China, in 2024. His research interests include LLMs interpretability, reliable evaluation and social network analysis.
\end{IEEEbiography}

\vskip -2\baselineskip plus -1fil

\begin{IEEEbiography}[{\includegraphics[width=1in,height=1.25in,clip,keepaspectratio]{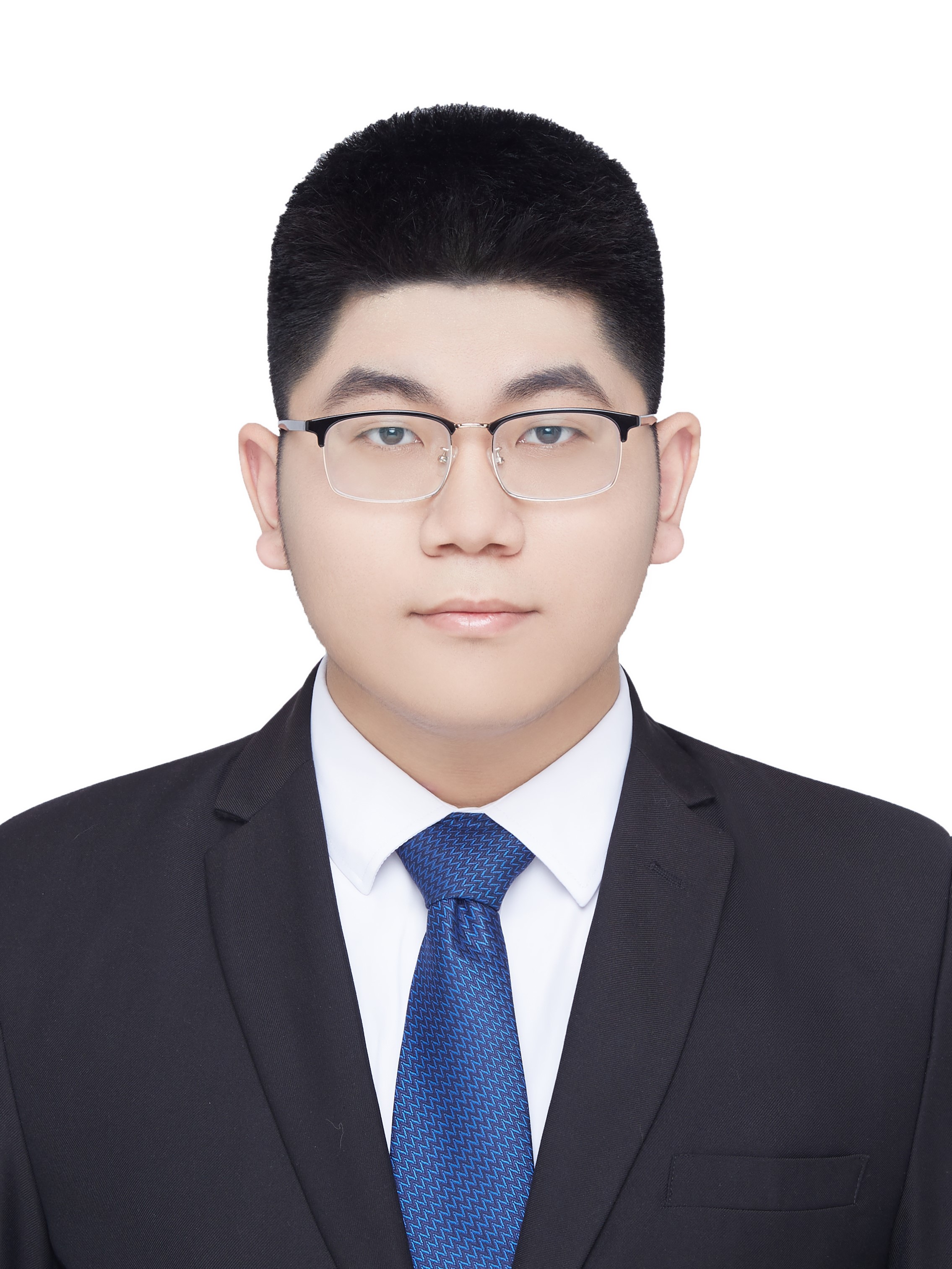}}]{Hanyu Wang}
is a Ph.D. student at the School of Information, Renmin University of China, under the supervision of Professor Xun Liang. His research areas include large language models, controllable text generation in large language models, and controlled decoding.
\end{IEEEbiography}

\vskip -2\baselineskip plus -1fil

\begin{IEEEbiography}[{\includegraphics[width=1in,height=1.25in,clip,keepaspectratio]{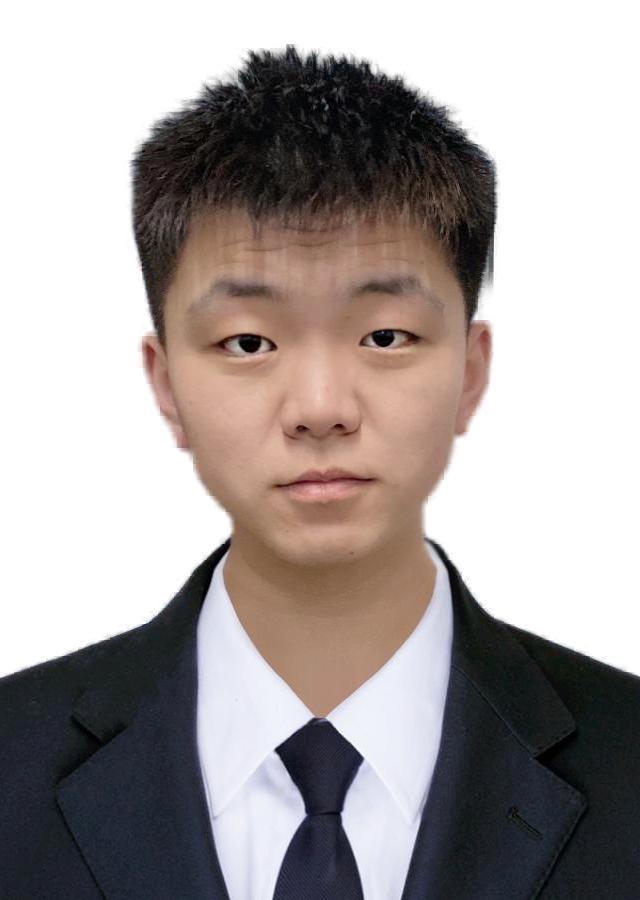}}]{Qingchen Yu}
is currently a research intern at the Large Language Model Center of the Institute for Advanced Algorithms Research in Shanghai. He is also a master's student at Shanghai University. His research interests include machine learning, LLM evaluation, and prompt engineering.
\end{IEEEbiography}

\vskip -2\baselineskip plus -1fil

\begin{IEEEbiography}[{\includegraphics[width=1in,height=1.25in,clip,keepaspectratio]{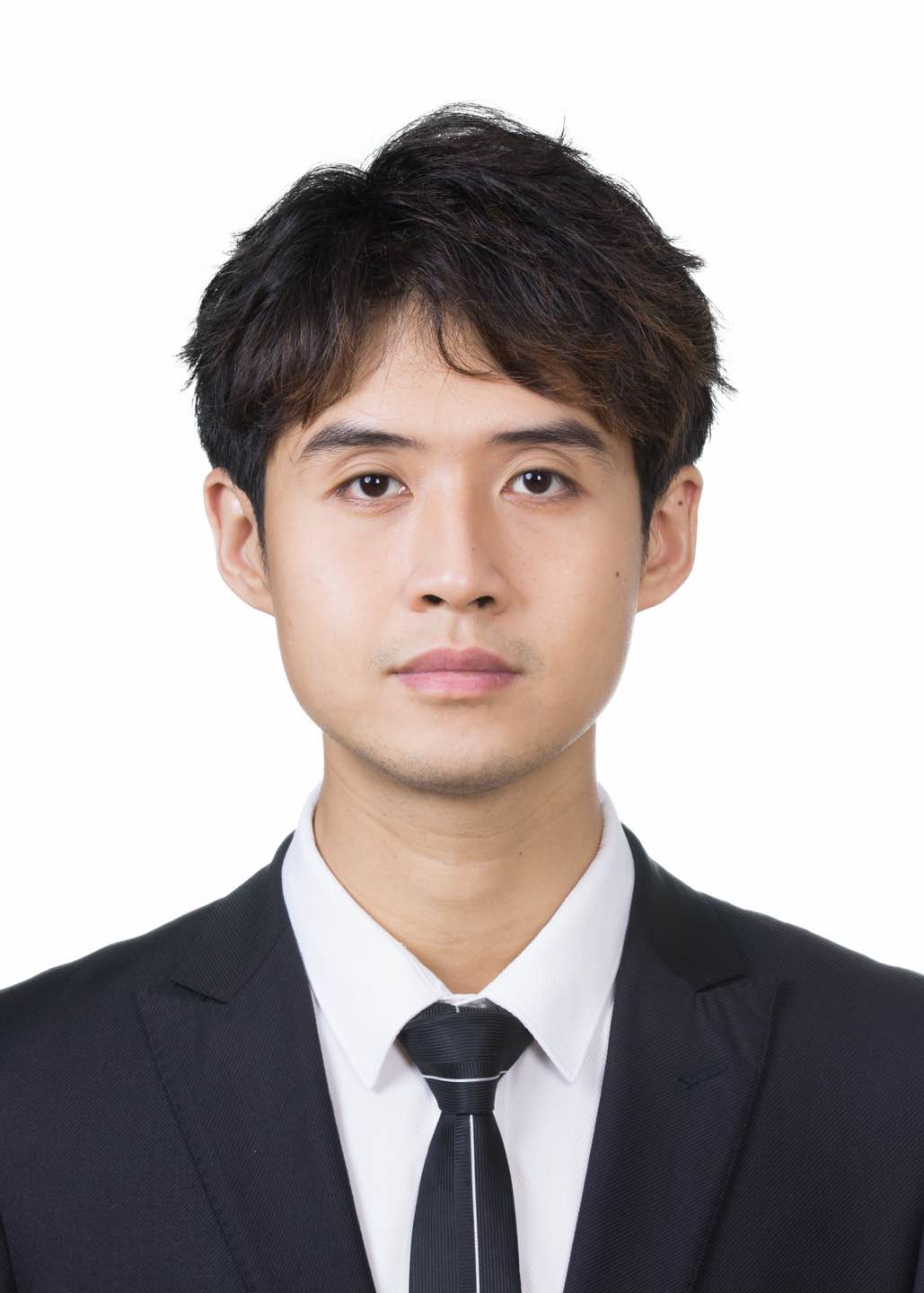}}]{Xunkai Li} is currently working toward the PhD degree with the school of Computer Science, Beijing Institute of Technology, advised by Prof. Rong-Hua Li. He received the BS degree in computer science from Shandong University in 2022. His research interest lies in Data-centric ML and Graph-ML within complex relational data and new learning paradigms. He has published 5+ papers in top DB/DM/AI conferences such as VLDB, WWW, AAAI as the first author.
\end{IEEEbiography}

 \vskip -2\baselineskip plus -1fil
 
\begin{IEEEbiography}[{\includegraphics[width=1in,height=1.25in,clip,keepaspectratio]{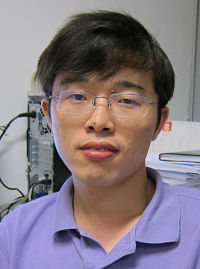}}]{Rong-Hua Li} received the Ph.D. degree in computer science from The Chinese University of Hong Kong, Hong Kong, in 2013. He is currently a Professor with the Beijing Institute of Technology, Beijing, China. His research interests include graph data management and mining, social network analysis, graph computation systems, and graph-based machine learning.
\end{IEEEbiography}

\vskip -2\baselineskip plus -1fil

\begin{IEEEbiography}[{\includegraphics[width=1in,height=1.25in,clip,keepaspectratio]{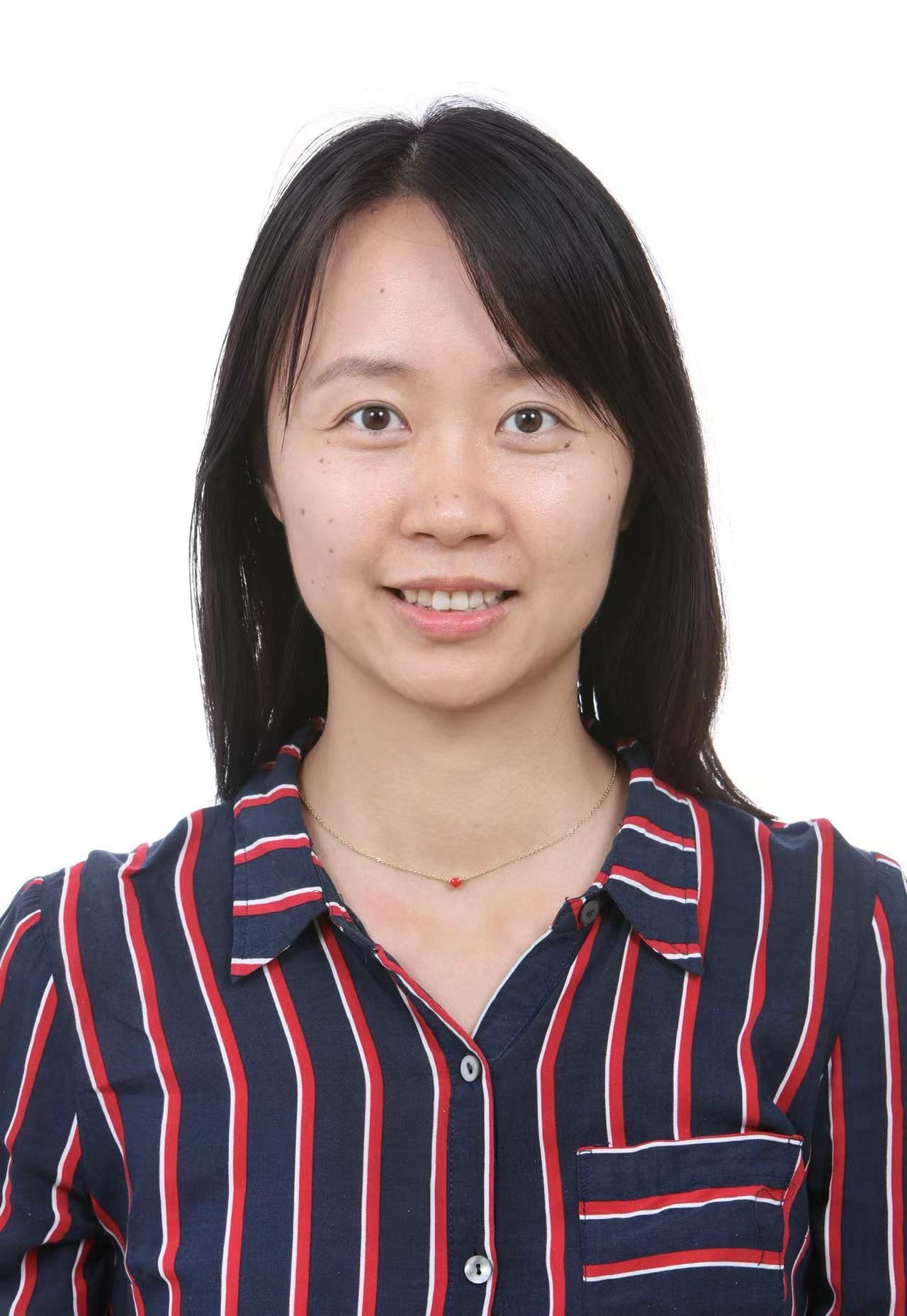}}]{Yi Wang} is a key member of the State Key Laboratory of Media Convergence Production Technology and Systems, a Senior Engineer in the Technology Department in Xinhua News Agency and one of the Xinhua News Agency 100 high-level talents. She has a long career engaged in news production and new technology innovation research. She is highly experienced in intelligent algorithm research and media integration, and expert in big data analysis and data mining.
\end{IEEEbiography}

\vskip -2\baselineskip plus -1fil

\begin{IEEEbiography}[{\includegraphics[width=1in,height=1.25in,clip,keepaspectratio]{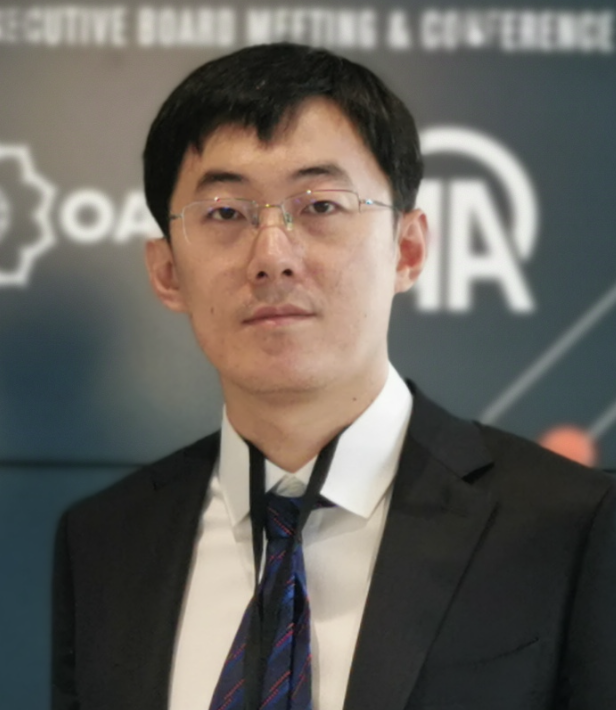}}]{Zhonghao Wang} is a Senior Algorithm Engineer at the State Key Laboratory of Media Convergence Production Technology and the AI Director at the Tech Bureau of Xinhua News Agency. He holds both a Bachelor's and a Master's degree from Shanghai Jiaotong University. He has previously served as an Algorithm Engineer in Alibaba's advertising department, where he specialized in developing interactive advertising algorithms. His primary interests lie in the application of algorithms and engineering in industry, with a particular focus on large-scale models and recommendation algorithms.
\end{IEEEbiography}

\vskip -2\baselineskip plus -1fil

\begin{IEEEbiography}[{\includegraphics[width=1in,height=1.25in,clip,keepaspectratio]{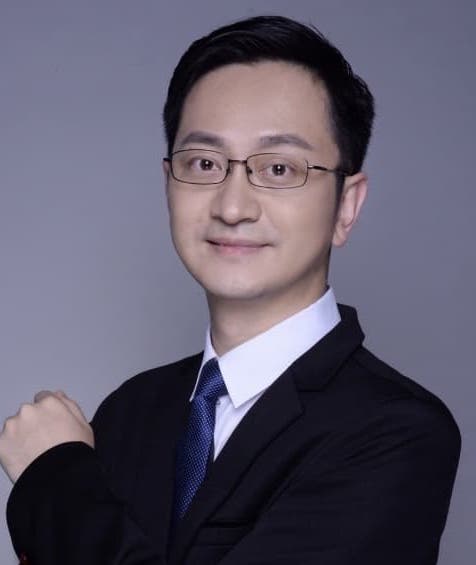}}]{Feiyu Xiong} is the Head of the Large Language Model Center of the Institute for Advanced Algorithms Research-Shanghai. He holds a Bachelor's degree from Huazhong University of Science and Technology and a Ph.D. from Drexel University. He has previously served as the Head of Data Intelligence for Alibaba's Business Middle Platform and the Head of the Data Platform for Taobao and Tmall Group. During his tenure at Alibaba, he was primarily responsible for the intelligent construction of systems related to core e-commerce transactions.
\end{IEEEbiography}

\vskip -2\baselineskip plus -1fil

\begin{IEEEbiography}[{\includegraphics[width=1in,height=1.25in,clip,keepaspectratio]{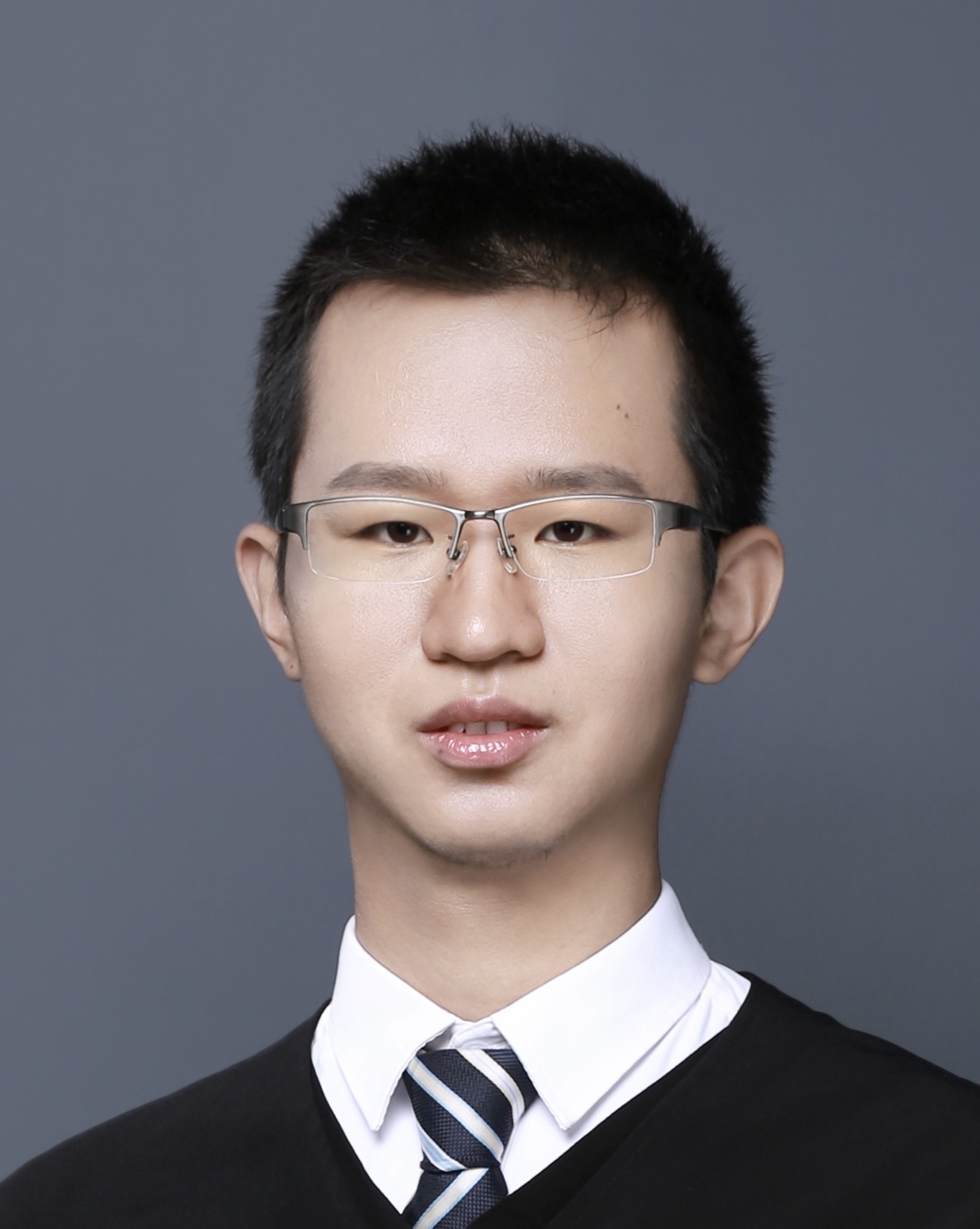}}]{Zhiyu Li} received his Ph.D. in Computer Science from the School of Information, Renmin University of China, in 2019. He is currently a Senior Researcher at the Large Language Model Center of the Institute for Advanced Algorithms Research-Shanghai. He has published over 30 papers in top-tier conferences and journals such as TKDE, KDD, and ACL. His current responsibilities include research and application implementation related to large language models. His research interests include model pre-training, model alignment, and hallucination optimization.
\end{IEEEbiography}

\vfill % This command ends the paragraph at the spot and adds the filling vertical space.

\end{document}